\documentclass{article}
\usepackage{graphicx}
\usepackage{enumitem}
\usepackage{amsmath,amssymb,amsthm}
\usepackage{bbm}
\usepackage{accents}
\usepackage{hyperref}
\usepackage{algorithm}
\usepackage{algorithmic}
\usepackage{multicol}
\usepackage{wrapfig}
\usepackage{multirow}

\usepackage{booktabs}

\usepackage{xspace}
\usepackage{subfigure}

  \providecommand{\yy}{\mathbf{y}}

  \providecommand{\mA}{\mathbf{A}}

  \providecommand{\mG}{\mathbf{G}}
  \providecommand{\mH}{\mathbf{H}}
  \providecommand{\mI}{\mathbf{I}}

  \providecommand{\mS}{\mathbf{S}}

  \providecommand{\mW}{\mathbf{W}}
  \providecommand{\mX}{\mathbf{X}}

  \providecommand{\cD}{\mathcal{D}}
  \providecommand{\cE}{\mathcal{E}}

    \providecommand{\cI}{\mathcal{I}}

  \providecommand{\cL}{\mathcal{L}}

  \providecommand{\cO}{\mathcal{O}}

  \providecommand{\cS}{\mathcal{S}}

  \providecommand{\cW}{\mathcal{W}}

\def\leref#1{Lemma~\ref{#1}}
\def\figref#1{Fig.~\ref{#1}}
\newtheorem{lemma}{Lemma}
\newtheorem{theorem}{Theorem}

\def\secref#1{Section~\ref{#1}}
\def\leref#1{Lemma~\ref{#1}}

\def\thref#1{Theorem~\ref{#1}}
\def\remref#1{Remark~\ref{#1}}

\def\figref#1{Figure~\ref{#1}}
\def\tabref#1{Table~\ref{#1}}
\def\algref#1{Algorithm~\ref{#1}}
\def\appref#1{Appendix~\ref{#1}}
\def\asref#1{A\ref{#1}}
\def\bydef{\triangleq}

\newcommand{\R}{\mathbb{R}}
\newcommand{\abs}[1]{\left\lvert #1\right\rvert}
\newcommand{\norm}[1]{\left\lVert #1\right\rVert}

\DeclareMathOperator{\E}{\mathbb{E}}

\newcommand{\lin}[1]{\ensuremath \left\langle #1 \right\rangle}

\newtheorem{assumption}{Assumption}

\theoremstyle{remark}
\newtheorem{remark}{Remark}
\usepackage[]{color-edits}

\usepackage{hyperref}
\usepackage{url}
\usepackage{enumitem}
\addauthor{mh}{blue}
\addauthor{xw}{cyan}
\addauthor{JC}{magenta}

\newcommand{\comment}[1]{}

\newcommand{\ourmethod}{GLASU\xspace}
\title{{GLASU: A Communication-Efficient Algorithm for Federated Learning with Vertically Distributed Graph Data}}
\author{Xinwei Zhang$^\dag$\thanks{Done at MIT-IBM Watson AI Lab.}, Mingyi Hong$^\dag$, and Jie Chen$^\ddag$\\
$^\dag$ Department of Electrical and Computer Engineering, University of Minnesota\\
$^\ddag$ MIT-IBM Watson AI Lab, IBM Research\\
$^\dag$\texttt{zhan6234,mhong@umn.edu}, $^\ddag$\texttt{chenjie@us.ibm.com}
}

\newcommand{\citep}[1]{\cite{#1}}
\newcommand{\citet}[1]{\cite{#1}}
\begin{document}
\maketitle
\setlist{nosep}
\begin{abstract}
Vertical federated learning (VFL) is a distributed learning paradigm, where computing clients collectively train a model based on the partial features of the same set of samples they possess. Current research on VFL focuses on the case when samples are independent, but it rarely addresses an emerging scenario when samples are interrelated through a graph. In this work, we train a graph neural network (GNN) through VFL, where each client owns a part of the node features and a different edge set. This data scenario incurs a significant communication overhead, not only because of the handling of distributed features but also due to neighborhood aggregation in a GNN. Moreover, the training analysis is faced with a challenge caused by the biased stochastic gradients. We propose a model-splitting method that splits a backbone GNN across the clients and the server and a communication-efficient algorithm, GLASU, to train such a model. GLASU adopts lazy aggregation and stale updates to skip communication in neighborhood aggregation and in model updates, respectively, greatly reducing communication while enjoying convergence guarantees. We conduct extensive numerical experiments on real-world datasets, showing that GLASU effectively trains a GNN that matches the accuracy of centralized training, while using only a fraction of the time due to communication saving.
\end{abstract}

\section{Introduction}\label{sec:intro}
Vertical federated learning (VFL) is a newly developed machine learning scenario in distributed optimization, where clients share data with the same sample identity but each client possesses only a subset of the features for each sample. The goal is for the clients to collaboratively learn a model based on all features. Such a scenario appears in many applications, including healthcare, finance, and recommendation systems.

Most of the current VFL solutions~\citep{chen2020vafl, liu2022fedbcd} treat the case where samples are independent, but omit their relational structure. However, the pairwise relationship between samples emerges in many occasions and it can be crucial in several learning scenarios, including the low-labeling-rate scenario in semi-supervised learning and the no-labeling scenario in self-supervised learning.

\begin{figure}[t]
  \centering
  \includegraphics[width=0.6\linewidth]{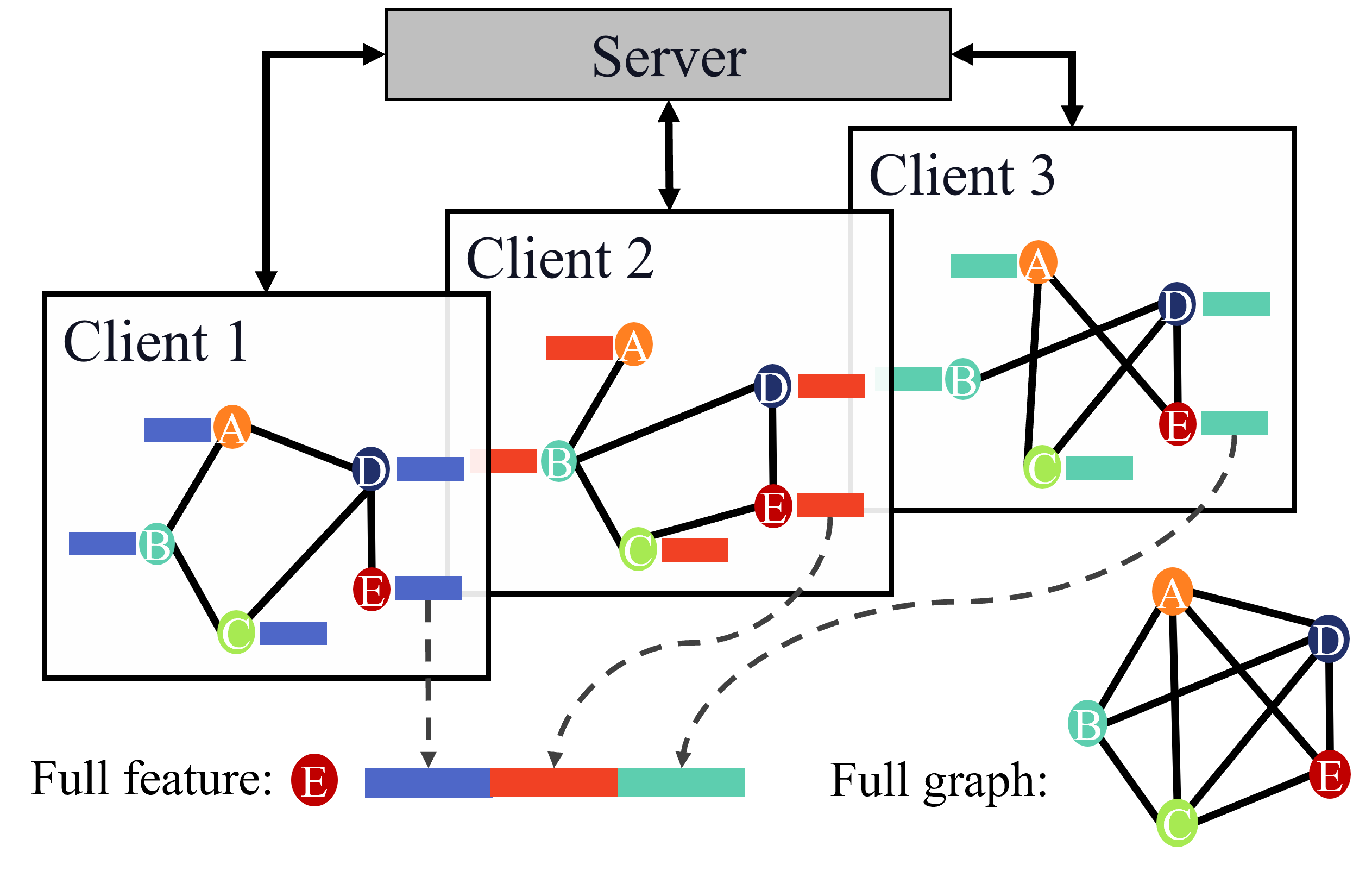}
  \vspace{-0.35cm}
  \caption{Data isolation of vertically distributed graph-structured data over three clients.}
  \label{fig:data}
  \vspace{-0.25cm}
\end{figure}

Consider, for example, a company that offers news recommendations to its subscribed users. Several departments may be maintaining a separate user graph in their own compute infrastructure: a professional network where users are connected through occupational ties; a personal network where users are connected through personal life interactions; a follower network where a user is a follower of another on social media, etc. Further, the user data in each graph may contain different features (e.g., occupation related, life related, and interest related, respectively). To offer personal recommendations, the company sets up a server that communicates with each client (each department's computer), to train a model that predicts multiple labels for each user without revealing each client's raw local data. See \figref{fig:data} for an illustration.

One of the most effective machine learning models for such a prediction task is graph neural networks (GNNs)~\citep{kipf2016semi,hamilton2017inductive,chen2018fastgcn,velickovic2018graph,chen2020simple}. This model performs neighborhood aggregation in every feature transformation layer, such that the prediction of a graph node is based on not only the information of this node but also that of its neighbors.

VFL on graph-structured data is not as well studied as that on other data, in part because of the challenges incurred by an enormous amount of communication. The communication overhead comes not only from the aggregation of the partial features/representations of a datum, but also from the neighborhood aggregation unique to GNNs. That is, communication occurs in each layer of the neural network, so that the latest representation of a neighboring node can be used to update the representation of the center node. One solution to reduce communication is that each client uses a local GNN to extract node representations from its own graph and the server aggregates these representations to make predictions~\citep{zhou2020vertically}. The drawback of this method is that the partial features of a node outside one client's neighborhood are not used, even if this node appears in another client's neighborhood. Another solution is to simulate centralized training: intermediate representations of each node are aggregated by the server, from where neighborhood aggregation is performed~\citep{ni2021vertical}. This method suffers the communication overhead incurred in each layer computation.

In this work, we propose \ourmethod for communication-efficient VFL on graph data. The GNN model is split across the clients and the server, such that the clients can use a majority of existing GNNs as the backbone, while the server contains no model parameters. The server only aggregates and disseminates processed data (e.g., node embeddings) with the clients. The communication frequency between the clients and the server is mitigated through \emph{lazy aggregation and stale updates} (hence the name of the method). For an $L$-layer GNN, \ourmethod communicates partial node representations only in $K$ layers and in every other $Q$ iterations, enjoying the reduction of communication by a factor of $QL/K$. \ourmethod can be considered as a framework that encompasses several well-known models and algorithms as special cases, including the work of~\citet{liu2022fedbcd} when the graphs are absent, the work of~\citet{zhou2020vertically} when all aggregations but the final one are skipped ($K=1$), the work of~\citet{ni2021vertical} when no aggregations are skipped ($K=L$), and centralized training when only a single client exists.

With the enjoyable reduction in communication, another difficulty is the convergence analysis, which admits two challenges: the biased gradient caused by neighborhood sampling in training GNNs and the correlated updates due to the use of stale node representations. We conduct an analysis based on the error decomposition of the gradient, showing that the training admits a convergence rate of $\cO((TQ)^{-1})$, where $T$ is the number of training rounds, each of which contains $Q$ iterations.

We summarize the main contributions of this work below:
\begin{enumerate}[leftmargin=*]
\item Model design: We propose a flexible, federated GNN architecture that is compatible with a majority of existing GNN models. 

\item Algorithm design: We propose the communication-efficient \ourmethod algorithm to train the model. Therein, lazy aggregation saves communication for each joint inference round, through skipping some aggregation layers in the GNN; while stale updates further save communication by allowing the clients to use stale global information for multiple local model updates.

\item Theoretical analysis: We provide theoretical convergence analysis for \ourmethod by addressing the challenges of biased stochastic gradient estimation caused by neighborhood sampling and correlated update steps caused by using stale global information. {To the best of our knowledge, this is the first convergence analysis for federated learning with graph data.} 

\item Numerical results: We conduct extensive experiments on seven datasets, together with ablation studies, to demonstrate that \ourmethod can achieve a comparable performance as the centralized model on multiple datasets and multiple GNN backbones, and that \ourmethod effectively saves communication and reduces training time. 
\end{enumerate}

\section{Problem, Background, and Related Works}

\subsection{Problem Setup}
Consider $M$ clients, indexed by $m = 1,\ldots,M$, each of which holds a part of a graph with the node feature matrix $\mX\in\R^{N \times d}$ and the edge set $\cE$. Here, $N$ is the number of nodes in the graph and $d$ is the feature dimension. We assume that each client has the same node set and the same set of training labels, $\yy$, but a different edge set $\cE_m$ and a non-overlapping node feature matrix $\mX_m\in\R^{N\times d_m}$, such that $\cE = \bigcup_{m=1}^M\cE_m$, $\mX=[\mX_1,\ldots,\mX_M]$, and $d=\sum_{m=1}^M d_m$. We denote the client dataset as $\cD_m = \{\mX_m, \cE_m, \yy\}$ and the full dataset as $\cD = \{\mX, \cE, \yy\}$. The task is for the clients to collaboratively infer the labels of nodes in the test set.

\subsection{Graph Convolutional Network}\label{sec:GCN}
The graph convolution network (GCN)~\citep{kipf2016semi} is a typical example of the family of GNNs. Inside GCN, a graph convolution layer reads
\begin{equation}\label{eqn:GCN}
\mH[l+1] = \sigma\Big(\mA(\cE)\cdot\mH[l]\cdot\mW[l]\Big),
\end{equation}
where $\sigma(\cdot)$ denotes the point-wise nonlinear activation function, $\mA(\cE)\in \R^{N\times N}$ denotes the adjacency matrix defined by the edge set $\cE$ with proper normalization, $\mH[l] \in \R^{N\times d[l]}$ denotes the node representation matrix at layer $l$, and $\mW[l] \in \R^{d[l]\times d[l+1]}$ denotes the weight matrix at the same layer.
The initial node representation matrix $\mH[0]=\mX$. The classifier is denoted as $\hat{\yy} = f(\mH[L],\mW[L])$ with weight matrix $\mW[L]$ and the loss function is denoted as $\ell(\yy, \hat{\yy})$. Therefore, the overall model parameter is $\mW = \{\mW[0],\dots,\mW[L-1],\mW[L]\}$.

Mini-batch training of GCN (and GNNs in general) faces a scalability challenge, because computing one or a few rows of $\mH[L]$ (i.e., the representations of a mini-batch) requires more and more rows of $\mH[L-1]$, $\mH[L-2]$, \ldots recursively, in light of the multiplication with $\mA(\cE)$ in~\eqref{eqn:GCN}. This is known as the \emph{explosive neighborhood problem} unique to graph-structured data. Several sampling strategies were proposed in the past to mitigate the explosion; in this work, we adopt the layer-wise sampling proposed by FastGCN~\citep{chen2018fastgcn}. Starting from the output layer $L$, which is associated with a mini-batch of training nodes, $\cS[L]$, we iterate over the layers backward such that at layer $l$, we sample a subset of neighbors for $\cS[l+1]$, namely $\cS[l]$. In doing so, at each layer, we form a bipartite graph with edge set $\cE[l] = \{(i,j)\vert i\in \cS[l+1], j\in\cS[l]\}$. Then, each graph convolution layer becomes
\begin{equation}\label{eqn:FastGCN}
\mH[l+1][\cS[l+1]] = \sigma\Big(\mA(\cE[l])\cdot\mH[l][\cS[l]]\cdot\mW[l]\Big),
\end{equation}
where $\mA(\cE[l])\in \R^{\abs{\cS[l+1]}\times\abs{\cS[l]}}$ is a properly scaled submatrix of $\mA(\cE)$ and $\mH[l][\cS[l]]$ denotes the rows of $\mH[l]$ corresponding to $\cS[l]$.

\subsection{Related Works}
\textbf{Vertical federated learning} is a learning paradigm where the features of the data are distributed across clients, who collaborate to train a model that incorporate all features~\citep{liu2022fedbcd, chen2020vafl,romanini2021pyvertical,yang2019parallel,gu2021privacy,yang2019quasi,xu2021fedv}. Thus, the global model is split among clients and the key challenge is the heavy communication costs on exchanging partial sample information for computing the losses and the gradients for each sample. Most works consider simple models (e.g., linear) because complex models incur multiple rounds of communication for prediction.

\textbf{Federated learning with graphs} includes four scenarios. The \emph{graph-level} scenario is \emph{horizontal}, where each client possesses a collection of graphs and all clients collaborate to train a unified model~\citep{zhang2021federated,he2021fedgraphnn,bayram2021federated,xie2021federated}. The task is to predict graph properties (such as molecular properties).

The \emph{subgraph-level} scenario could be either \emph{vertical} or \emph{horizontal}. In the vertical scenario, each client holds a part of the node features, a part of the whole model, and additionally a subgraph of the global graph~\citep{zhou2020vertically,ni2021vertical}. The clients aim to collaboratively train a global model (combined from those of each client) to predict node properties (such as the category of a paper in a citation network). Our work addresses this scenario.

The \emph{subgraph-level, horizontal} scenario, on the other hand, considers training a GNN for node property prediction in a \emph{distributed} manner: a graph is partitioned and each client holds one partition~\citep{zhang2021subgraph,wu2021fedgnn,chen2022fedgraph,yao2022fedgcn}. A challenge to address is the aggregation of information along edges crossing different clients. This scenario differs from the vertical scenario in that features are not partitioned among clients and the graph partitions do not overlap.

The fourth scenario is \emph{node-level}: the clients are connected by a graph and thus each of them is treated as a node. In other words, the clients, rather than the data, are graph-structured. It is akin to \emph{decentralized learning}, where clients communicate to each other via the graph to train a unified model~\citep{lalitha2019peer,meng2021cross,caldarola2021cluster,rizk2021graph}.

Due to the space limitation, please see \appref{app:related} for in-depth discussions of the related works.

\section{Proposed Approach}\label{sec:approach}
In this section, we present the proposed model and the training algorithm \ourmethod for federated learning on vertically distributed graph data. The neighborhood aggregation in GNNs poses communication challenges distinct from conventional VFL. To mitigate this challenge, we propose lazy aggregation and stale updates to effectively reduce the communication between the clients and the server, while maintaining comparable prediction performance as centralized models. For notational simplicity, we present the approach by using the full-graph notation~\eqref{eqn:GCN} but note that the implementation involves neighborhood sampling, where a more precise notation should follow~\eqref{eqn:FastGCN}, and that one can easily change the backbone from GCN to other GNNs.

\begin{figure*}[t]
  \begin{minipage}[t]{.65\linewidth}
    \centering
    \includegraphics[width=\linewidth]{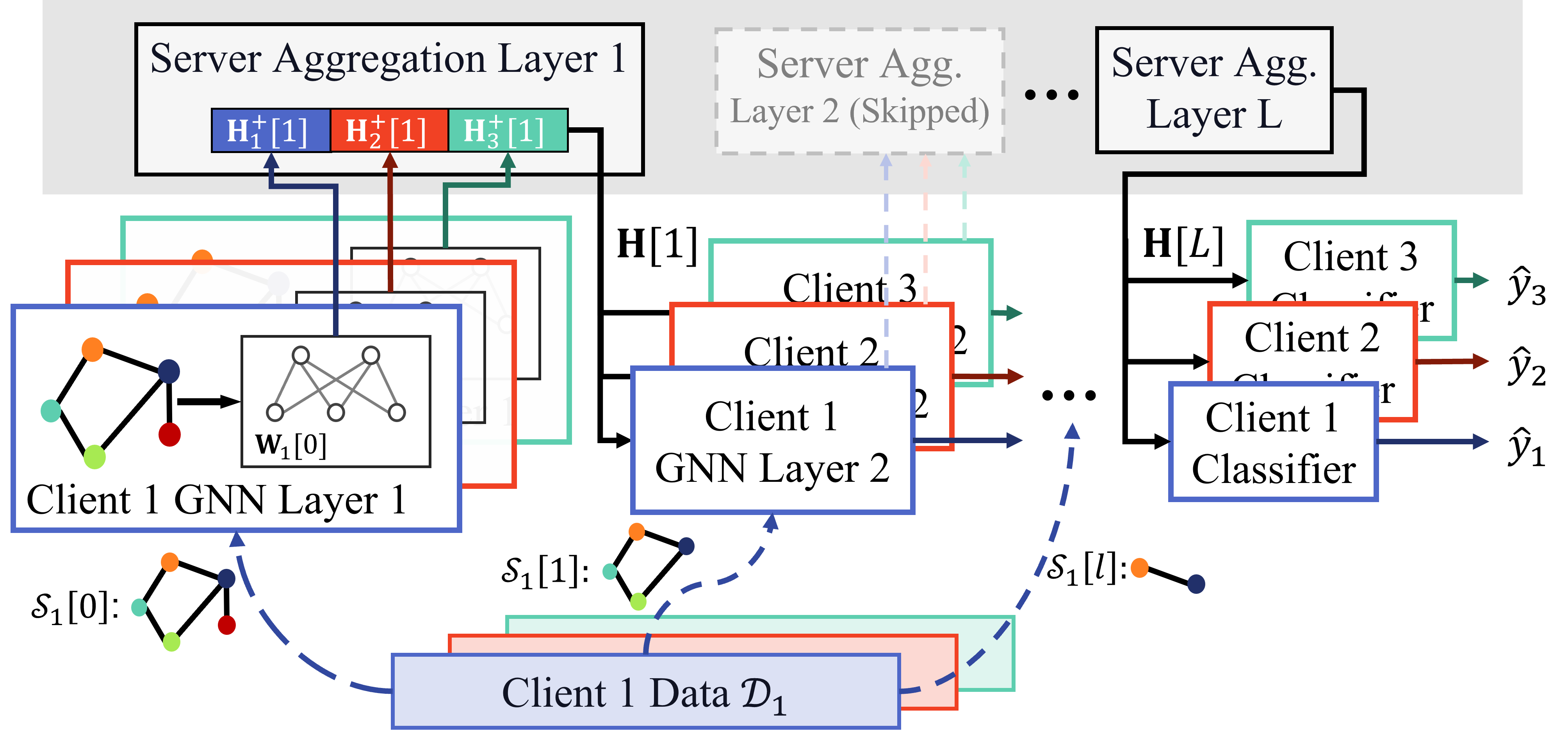}
  \end{minipage}%
  \hspace*{\fill}
  \begin{minipage}[b]{.32\linewidth}
    \caption{Illustration of the split model on $M=3$ clients with lazy aggregation. In the model, the second server aggregation layer is skipped and the graph size used by each layer gradually decreases, due to neighborhood aggregation (inverse of neighborhood sampling).}
    \vskip 10pt
    \label{fig:model}
  \end{minipage}
  \vskip -0.1in
\end{figure*}

\subsection{GNN Model Splitting}\label{sec:MS}
We split the GNN model among the clients and the server, approximating a centralized model. Specifically, each GNN layer contains two sub-layers: the client GNN sub-layer and the server aggregation sub-layer. At the $l$-th layer, each client computes the local feature matrix
\[
\mH^+_m[l] = \sigma\Big(\mA(\cE_m)\cdot\mH_m[l]\cdot\mW_m[l]\Big)
\]
with the local weight matrix $\mW_m[l]$ and the local graph $\cE_m$, where we use the superscript $^+$ to denote local representations before aggregation. Then, the server aggregates the clients' representations and outputs $\mH[l+1]$ as
\[
\mH[l+1] = \mathrm{Agg}(\mH^+_1[l],\dots,\mH^+_M[l]),
\]
where $\mathrm{Agg}(\cdot)$ is an aggregation function. In this paper, we only consider parameter-free aggregations, including averaging and concatenation. The server broadcasts the aggregated $\mH[l+1]$ to the clients so that computation proceeds to the next layer. In the final layer, each client computes a prediction. This layer is the same among clients because they receive the same $\mH[L]$.

The two aggregation operations of our choice render a rather simple implementation of the server. They bring in two advantages: parameter-free and memory-less. Since the operations do not contain any learnable parameters, the server does not need to perform gradient computations. Moreover, in the backward pass, these operations do not require data from the forward pass to back-propagate the gradients (memory-less). Specifically, for averaging, the server back-propagates $\frac{1}{M}\nabla_{\mH[l+1]}\cL$ to each client, where $\cL$ denotes the loss; while for concatenation, the server back-propagates the corresponding block of $\nabla_{\mH[l+1]}\cL$.

We illustrate in \figref{fig:model} the split of each GNN layer among the clients and the server. Note the difference of our approach from existing approaches. Our model splitting resembles federated split learning (SplitFed)~\citep{thapa2022splitfed}; but in SplitFed, each client can collaborate with the server to perform inference or model updates without accessing information from other clients, whereas in our case, all clients collectively perform the job. Our approach also differs from conventional VFL that splits the local feature processing and the final classifier among the clients and the server respectively, such that each model update requires a single U-shape communication~\citep{chen2020vafl}. In our case, due to the graph structure, each GNN layer contains one client-server interaction and the number of interactions is equal to the number of GNN layers (we will relax this in the following subsection).

\subsection{Lazy Aggregation}\label{sec:LLA}
The development in the preceding subsection approximates a centralized model, but it is not communication friendly because each layer requires one round of client-server communication. We propose two communication-saving strategies in this subsection and the next. We first consider \emph{lazy aggregation}, which skips aggregation in certain layers.

Instead of performing server aggregation at each layer, we specify a subset of $K$ indices, $\cI=\{l_1,\dots,l_K\} \subset [L]$, such that aggregation is performed only at these layers. That is, at a layer $l\in \cI$, the server performs aggregation and broadcasts the aggregated representations to the clients, serving as the input to the next layer:
\[
\mH_m[l+1] = \mH[l+1];
\]
while at a layer $l \notin \cI$, each client uses the local representations as the input to the next layer:
\[
\mH_m[l+1] = \mH^+_m[l].
\]
By doing so, the amount of communication is reduced from $\cO(L)$ to $\cO(K)$.

There are subtleties caused by neighborhood sampling, similar to those faced by FastGCN (see \secref{sec:GCN}). First, it requires additional rounds of communication to synchronize the sample indices, because whenever server aggregation is performed, it must be done on the same set of sampled nodes across clients. Hence, in the additional communication rounds, the server takes the union of the clients' index sets $\cS_m[l_k]$ and broadcasts $\cS[l_k] = \bigcup_{m=1}^M \cS_m[l_k]$ to the clients. Second, when server aggregation is skipped at a layer $l\notin \cI$, each client can use its own set of sampled nodes, $\cS_m[l]$, which may differ from each other. Such a procedure is more flexible than conventional VFL where sample features are generally processed synchronously. The sampling procedure is summarized in \algref{alg:sample} in \appref{app:algorithm}.

\subsection{Stale Updates}\label{sec:SU}
To further reduce communication, we consider \emph{stale updates}, which skip aggregation in certain iterations and use stale node representations to perform model updates. The key idea is to use the same mini-batch, including the sampled neighbors at each layer, for training $Q$ iterations. In every other $Q$ iterations, the clients store the aggregated representations at the server aggregation layers. Then, in the subsequent iterations, every server aggregation is replaced by a local aggregation between a client's up-to-date node representations and other clients' stale node representations. By doing so, the clients and the server only need to communicate once in every $Q$ iterations.

Specifically, let a round of training contain $Q$ iterations and use $t$ to index the rounds. At the beginning of each round, the clients and the server jointly decide the set of nodes used for training at each layer. Then, they perform a joint inference on the representations $\mH_m^{t,+}[l]$ at every layer $l\in\cI$. Each client $m$ will store the ``all but $m$'' representation $\mH_{-m}^t[l+1]$ through extracting such information from the aggregated representations $\mH_m^t[l+1]$:
\[
\mH^t_{-m}[l+1]= \mathrm{Extract}(\mH^t_{m}[l+1], \mH^{t,+}_{m}[l]).
\]
For example, when the server aggregation is averaging, the extraction is
\[
\mathrm{Extract}(\mH^t_{m}[l+1], \mH^{t,+}_{m}[l]) = \mH^t_{m}[l+1] - \frac{1}{M}\mH^{t,+}_{m}[l],
\]
Afterward, the clients perform $Q$ iterations of model updates, indexed by $q = 0,\dots, Q-1$, on the local parameters $\mW_m^{t,q}$ in parallel, using the stored aggregated information $\mH_{-m}^t[l+1]$ to perform local computation, replacing server aggregation. The name ``stale updates'' comes from the fact that $\mH_{-m}^t[l+1]$ is computed by using stale model parameters $\{\mW_{m'}^{t,0}\}_{m'\neq m}$ at all iterations $q\ne0$. The extraction and the local updates are summarized in \algref{alg:local_update:1} and \algref{alg:local_update:2}, respectively, in \appref{app:algorithm}.

\begingroup
\setlength{\textfloatsep}{6pt}
\begin{algorithm}[t]
  \begin{algorithmic}[1]
    \FOR{$t = 0, \dots, T$}
    \STATE {\bf Server/Client} (\algref{alg:sample}): Sample $\{\cS_m^t[l]\}^L_{l=0}$.
    \STATE {\bf Client}: $\mW_m^{t,0} = \begin{cases}\mW_m^{t-1,Q}, & t>0\\ \mW_m^0, &t = 0\end{cases}.$
      \STATE {\bf Server/Client} (\algref{alg:local_update:1}): $\{\mH^t_{-m}[l+1]\}_{l\in\cI} = ${\bf JointInference}$(\mW_m^{t,0}, \cD_m, \{\cS_m^t[l]\}^L_{l=0}).$
      \FOR {$q = 0,\dots, Q-1$}
      \STATE {\bf Client} (\algref{alg:local_update:2}): $\mW_m^{t,q+1}=$ {\bf LocalUpdate} $(\mW_m^{t,q}, \cD_m, \{\cS_m^t[l]\}^L_{l=0},\{\mH^t_{-m}[l+1]\}_{l\in\cI}).$
      \ENDFOR
      \ENDFOR
      \STATE \textbf{Output:} $\{\mW_m^{T,Q}\}^{M}_{m=1}$
  \end{algorithmic}
  \caption{Training Procedure. All referenced algorithms are detailed in \appref{app:algorithm}.}
  \label{alg:train}
\end{algorithm}

\subsection{Summary}
The overall training procedure is summarized in \algref{alg:train}. For communication savings, lazy aggregation brings in a factor of $L/K$ and stale updates bring in a factor of $Q$. Therefore, the overall saving factor is $QL/K$.

Note that the algorithm assumes that all clients have the training labels. If the labels can be held by only one client (say, A), a slight modification by broadcasting the gradient with respect to the final-layer output possessed by A, suffices. See \appref{app:algorithm_v2} for details.

\subsection{Special Cases}\label{sec:special}
It is interesting to note that \ourmethod encompasses several well-known methods as special cases.

\textbf{Conventional VFL.} VFL algorithms can be viewed as a special case of \ourmethod, where $\mA(\cE_m) = \mI$ for all $m$. In this case, no neighborhood sampling is needed and \ourmethod reduces to~\citet{liu2022fedbcd}.

\textbf{Existing VFL algorithms for graphs.} The model of~\citet{zhou2020vertically} is a special case of \ourmethod, with $K=1$; i.e., no communication is performed except the final prediction layer. In this case, the clients omit the connections absent in the self subgraph but present in other clients' subgraphs. The model of~\citet{ni2021vertical} is also a special case of \ourmethod, with $K = L$. This case requires communication at all layers and is less efficient.

\textbf{Centralized GNNs.} When there is a single client ($M=1$), our setting is the same as centralized GNN training. Specifically, by letting $K=L$ and properly choosing the server aggregation function $\mathrm{Agg}(\cdot)$, our split model can achieve the same performance as a centralized GNN model. Of course, using lazy aggregation ($K\ne L$) and choosing the server aggregation function as concatenation or averaging will make the split model different from a centralized GNN.

\endgroup
\subsection{Privacy}
\ourmethod enables privacy protection because it is compatible with existing privacy-preserving approaches.

\textbf{Secure Aggregation (SA)}~\citep{bonawitz2017practical,hardy2017private} is a form of secure multi-party computation approach used for aggregating information from a group of clients, without revealing the information of any individual. This can be achieved by homomorphic encryption~\citep{li2010secure,hardy2017private}. In our case, when the server aggregation is averaging, homomorphic encryption can be directly applied.

\textbf{Differential Privacy (DP)}~\citep{wei2020federated} is a probabilistic protection approach. By injecting stochasticity into the local outputs, this approach guarantees that an attacker cannot distinguish the sample from the dataset up to a certain probability. DP can be applied either solely or in combination with SA to our algorithm in the server-client communication, to offer privacy protection for the client data.

\section{Convergence Analysis}\label{sec:theory:converge}
In this section, we analyze the convergence behavior of \ourmethod under lazy aggregation and stale updates. To start the analysis, denote by $\cS^t = \{\cS^t_m[l]\}^{L,M}_{l=1,m=1}$ the samples used at round $t$ (which include all sampled nodes at different layers and clients); by $S = \abs{\cS^t_m[L]}$ the batch size; and by $\cL(\mW;\cS)$ the training objective, which is evaluated at the overall set of model parameters across clients, $\mW = \{\mW_m\}^M_{m=1}$, and a batch of samples, $\cS$.

A few assumptions are needed (see Appendix~\ref{app:theory:as} for formal statements). \textbf{A1}: The loss function $\ell$ is $G_\ell$-smooth with $L_\ell$-Lipschitz gradient; and a client's prediction function $f_m$ is $G_f$-smooth with $L_f$-Lipschitz gradient.
\textbf{A2}: The training objective $\cL(\mW;\cD)$ is bounded below by a finite constant $\cL^\star$.
\textbf{A3}: The samples $\cS^t$ are uniformly sampled from the neighbor set in each layer.

\begin{theorem}\label{thm:convergence}
  Under assumptions A1--A3, by running \algref{alg:train} with constant step size $\eta \leq C_0^{-1}\cdot (1+2Q^2M)^{-1}$, with probability at least $p = 1-\delta$, the averaged squared gradient norm is bounded by:
  \begin{equation*}
    \frac{1}{TQ}\sum^{T-1}_{t=0}\sum^{Q-1}_{q=0}\E\norm{\nabla \cL(\mW^{t,q};\cD)}^2 \leq  \frac{2\Delta_{\cL}}{\eta TQ}+\frac{28\eta M\cdot\left(C_0 + \sqrt{M+1}Q\right)}{3}\sigma,
  \end{equation*}
  where $\Delta_{\cL}=\cL(\mW^{0,0})-\cL^\star$, $C_0 = G_\ell L_f + L_\ell G_f^2$, and $\sigma>0$ is a function of $\log(TQ/\delta), L_f, L_g, G_f$ and $G_g$.
\end{theorem}

\begin{remark}
There are two key challenges in the analysis. (1) Owing to neighborhood sampling, the stochastic gradient is biased (i.e., $\E_\cS\nabla \cL(\mW; \cS) \neq \nabla \cL(\mW; \cD)$). (2) The stale updates in one communication round are correlated, as they use the same mini-batch and samples. Hence, the general unbiasedness and independence assumptions on the stochastic gradients in the analysis of SGD-type of algorithms do not apply. We borrow the technique by~\citet{ramezani2020gcn} to bound the error of the stochastic gradient through the bias-variance decomposition and extend the analysis by~\citet{liu2022fedbcd} for VFL with correlated updates to establish our proof. For details, see \appref{app:theory}.
\end{remark}

\begin{remark}\label{rem:stationary}
To better expose the convergence rate, assuming that $Q$ is upper bounded by $\frac{C_0}{\sqrt{M+1}}$, one may set $\eta = \sqrt{\frac{3\Delta_{\cL}}{28MC_0\sigma TQ}}$, such that
\[
  \frac{1}{TQ}\sum^{T-1}_{t=0}\sum^{Q-1}_{q=0}\E\norm{\nabla \cL(\mW^{t,q};\cD)}^2
  \leq 8\sqrt{\frac{7\Delta_{\cL}MC_0\sigma}{3TQ}}.
\]
Ignoring the logarithmic factor $\log(TQ/\delta)$ in $\sigma$, the above bound states that the squared gradient norm decreases as $\cO((TQ)^{-1})$. Note that this bound holds only when $T$ is sufficiently large, because the choice of $\eta$ must satisfy the condition of \thref{thm:convergence}.
\end{remark}

\begin{remark}\label{rem:QT}
Based on the preceding remark, we see that to achieve $\epsilon$-stationarity, the number of model updates is $QT = \cO(\frac{1}{\epsilon^2})$. That is, as long as $Q$ obeys the upper bound, running more local updates ($Q$) reduces the amount of communications ($T$). To the best of our knowledge, this is the first result for VFL on graph data.
\end{remark}

\begin{remark}
While we have analyzed the impact of stale updates ($Q$), lazy aggregation ($K$) does not play a role in convergence, because it does not affect model updates. Instead, it affects model accuracy in a manner similar to how changing a neural network impacts the prediction accuracy.
\end{remark}

\begin{remark}
If we consider the impact of the number of clients, the factor $M$ in the numerator of the bound indicates a slowdown when more clients participate training. Similar results are seen in FedBCD~\cite{liu2022fedbcd}, but therein one can use a large batch size $S$ to counter the slowdown. For graphs, however, $S$ does not appear in the bound because of the biased gradient estimation. Nevertheless, we note that unlike other federated scenarios, in VFL, $M$ is very small because it is limited by, e.g., the feature length.
\end{remark}

\section{Numerical Experiments}
In this section, we conduct numerical experiments on a variety of datasets and demonstrate the effectiveness of \ourmethod in training with distributed graph data. We first compare its performance with related methods, including those tackling a different assumption on the data distribution and communication pattern. Then, we examine the communication saving owing to the use of lazy aggregation and stale updates. We further showcase the flexibility of \ourmethod through demonstration with different GNN backbones and varying clients. The experiments are conducted on a distributed cluster with three Tesla V100 GPUs communicated through Ethernet.

\subsection{Datasets}
We use seven datasets (in three groups) with varying sizes and data distributions: the Planetoid collection~\citep{yang2016revisiting}, the HeriGraph collection~\citep{bai2022heri}, and the Reddit dataset~\citep{hamilton2017inductive}. Each dataset in the HeriGraph collection (Suzhou, Venice, and Amsterdam) contains data readily distributed: three subgraphs and more than three feature blocks for each node. Hence, we use three clients, each of which handles one subgraph and one feature block. For the other four datasets (Cora, PubMed, and CiteSeer in the Planetoid collection; and Reddit), each contains one single graph and thus we manually construct subgraphs through randomly sampling the edges and splitting the input features into non-overlapping blocks, so that each client handles one subgraph and one feature block. The dataset statistics are summarized in Table~\ref{tab:dataset} and more details are given in Appendix~\ref{app:exp:data}. 

\begin{table}[t]
  \centering
  \caption{Datasets. Each of the HeriGraph datasets (Suzhou, Venice, Amsterdam) contains three naturally formed subgraphs. For other datasets, each contains one single graph and each client holds a sampled subgraph of it.}
  \label{tab:dataset}
  \vskip 5pt
  \small
  \begin{tabular}{c|rrrc}
    \toprule
    Dataset & \# Nodes & \# Edges & \# Feat. & \# Class \\
    \midrule
    Cora & $2,708$ & $10,556$ & $1,433$ & $7$ \\
    PubMed & $19,717$ & $88,648$ & $500$ & $3$\\
    CiteSeer & $3,327$ & $9,104$ & $3,703$ & $6$\\
    \midrule
    Suzhou & $3,137$ & $916,496$ & $979$ & $9$ \\
    Venice & $2,951$ & $534,513$ & $979$ & $9$ \\
    Amsterdam & $3,727$ & $1,271,171$ & $979$ & $9$ \\
    \midrule
    Reddit & $232,965$ & $114,615,892$ & $602$ & $41$ \\
    \bottomrule
  \end{tabular}
  \vskip -0.1in
\end{table}

\subsection{Accuracy}
We compare \ourmethod with three training methods: (a) centralized training, where there is only a single client ($M=1$), which holds the whole dataset without any data distribution and communication; (b) standalone training~\citep{zhou2020vertically}, where each client trains a model with its local data only and they do not communicate; (c) simulated centralized training~\citep{ni2021vertical}, where each client possesses the full graph but only the partial features, so that it simulates centralized training through server aggregation in each GNN layer. Methods (b) and (c) are typical VFL baselines; they are also special cases of our method (see \secref{sec:special}). Except for centralized training, the number of clients $M=3$. The number of training rounds, $T$, and the learning rate $\eta$ are optimized through grid search. See Appendix~\ref{app:exp:hyperparam} for details.

We use GCNII~\citep{chen2020simple} as the backbone GNN. GCNII improves over GCN through including two skip connections, one with the current layer input and the other with the initial layer input.
We set the number of layers $L=4$ and the mini-batch size $S=16$. For neighborhood sampling, the sample size is three neighbors per node on average. We set $K=2$; i.e., lazy aggregation is performed in the middle and the last layer.

Table~\ref{tab:accuracy} reports the average classification accuracy of \ourmethod and the compared training methods, repeated five times. As expected, standalone training produces the worst results, because each client uses only local information and misses edges and node features present in other clients. The centralized training and its simulated version lead to similar performance, also as expected, because server aggregation (or its equivalence in centralized training) on each GNN layer takes effect. Our method \ourmethod, which skips half of the aggregations, yields prediction accuracy rather comparable with these two methods. Using stale updates ($Q=4$) is generally outperformed by no stale updates ($Q=1$), but occasionally it is better (see PubMed and Amsterdam). The gain in using lazy aggregation and stale updates occurs in timing, as will be demonstrated next.

\begin{table*}[t]
  \centering
  \caption{Test accuracy ($\%$). The compared algorithms are Centralized training (Cent.); Standalone training (StAl.); Simulated centralized training (Sim.); \ourmethod with no stale updates, i.e., $Q=1$ (\ourmethod-1); and \ourmethod with stale updates $Q=4$ (\ourmethod-4).}
  \label{tab:accuracy}
  \vskip 5pt
  \small
  \begin{tabular}{c|ccc|cc}
    \toprule
    Dataset      & Cent. & StAl. & Sim. & \ourmethod-1 & \ourmethod-4 \\
    \midrule
    Cora        & $80.9\pm0.6$ & $74.6\pm0.5$ & $80.1\pm1.2$ & $81.0\pm1.3$ & $80.3\pm1.2$ \\
    PubMed      & $84.9\pm0.6$ & $77.2\pm 0.5$  & $82.7\pm1.2$ & $82.3\pm1.6$ & $83.8\pm1.8$  \\  
    CiteSeer    & $70.2\pm0.8$  & $64.4\pm0.5$   & $70.0\pm1.2$& $70.0\pm1.7$  & $68.8\pm3.3$ \\
    \midrule
    Suzhou   & $94.3\pm0.3$   & $51.6\pm0.9$  & $93.5\pm0.6$    & $92.7\pm1.4$  & $90.4\pm0.8$   \\  
    Venice   & $95.7\pm 0.5$ & $33.5 \pm 2.1$    & $93.1\pm1.3$    & $92.2\pm0.6$  & $91.0\pm1.6$   \\ 
    Amsterdam    & $94.6\pm0.1$  & $59.8\pm1.0$    & $95.5\pm0.8$    & $93.1\pm0.8$    & $94.9\pm0.4$  \\ 
    \midrule
    Reddit       & $95.6\pm0.1$   & $87.3\pm0.3$    & $95.3\pm0.7$  & $95.7\pm0.6$  & $94.7\pm1.1$  \\ 
    \bottomrule
  \end{tabular}
  \vskip -0.1in
\end{table*}

\subsection{Communication Saving}
To further investigate how the two proposed techniques affect the model performance and save the communication, we conduct a study on (a) the lazy aggregation parameter $K$ and (b) the stale update parameter $Q$.

\textbf{Lazy aggregation:} We use a 4-layer GCNII as the backbone and set $K=1,2,4$. The aggregation layers are ``uniform'' across the model layers. That is, when $K=1$, server aggregation is performed on the last layer; when $K=2$, on the middle layer and the last layer; and when $K=4$, on all layers. The test accuracy and runtime are listed in Table~\ref{tab:abl:LA}. We observe that the runtime decreases drastically when using fewer and fewer aggregation layers: from $K=4$ to $K=1$, the reduction is $37.5\%$ for PubMed and $58.2\%$ for Amsterdam. The accuracy is comparable in all cases.

\begin{table}[t]
  \centering
  \caption{Test accuracy (\%), runtime (seconds), and saving in runtime (\%) under different numbers of lazy aggregation layers ($K =4,2,1$). The saving is with respect to $K=4$. Top: PubMed; bottom: Amsterdam.}
  \label{tab:abl:LA}
  \vskip 5pt
  \small
  \begin{tabular}{l | ccc}
    \toprule
    \# Layer            & $K=4$             & $K=2$             & $K=1$             \\
    \midrule
    Accuracy      & $82.5\pm1.0$      & $83.8\pm1.8$      & $82.2\pm0.7$      \\
    Runtime        &  $130\pm12$       & $96.6\pm9.9$      & $81.3\pm6.5$      \\
    Saving        & $-$               & $25.7$            & $37.5$            \\
    \bottomrule 
    \toprule
    \# Layer            & $K=4$             & $K=2$             & $K=1$             \\
    \midrule
    Accuracy      & $93.6\pm0.7$      & $94.9\pm0.4$      & $92.0\pm1.7$      \\
    Runtime        & $913\pm76$        & $544\pm44$        & $382\pm35$        \\
    Saving        & $-$               & $40.4$            & $58.2$            \\
    \bottomrule
  \end{tabular}
  \vskip -0.1in
\end{table}

\textbf{Stale updates:} We experiment with a few choices of $Q$: 2, 4, 8, and 16. We report the time to reach the same test accuracy threshold in Table~\ref{tab:abl:SU}. We see that stale updates help speed up training by using fewer communication rounds, corroborating \remref{rem:QT} of the theory in \secref{sec:theory:converge}. This trend occurs on the Amsterdam dataset even when taking $Q$ as large as 16. The trend is also noticeable on PubMed, but at some point ($Q=8$) it is reverted, likely because it gets harder and harder to reach the accuracy threshold. We speculate that the target 82\% can never be achieved at $Q=16$. This observation is consistent with \remref{rem:stationary} of the theory, requiring $Q$ to be upper bounded to claim $\cO((TQ)^{-1})$ convergence.

\begin{table}[t]
  \centering
  \caption{Test accuracy (\%) and runtime (seconds) under different numbers of stale updates ($Q =2,4,6,16$) for the same accuracy threshold. Top: PubMed (threshold: $82\%$); bottom: Amsterdam (threshold: $89\%$).}
  \label{tab:abl:SU}
  \vskip 5pt
  \small
  \begin{tabular}{l | cccc}
    \toprule
    \# Stale     & $Q = 2$           & $Q = 4$           & $Q=8$             & $Q=16$             \\
    \midrule
    Accuracy     & $82.5\pm1.6$      & $82.0\pm2.4$      & $82.1\pm0.3$      & N/A                \\
    Runtime       & $66.1\pm5.0$      & $43.8\pm4.0$      & $88.9\pm7.4$      & $>128$      \\
    \bottomrule
    \toprule
    \# Stale     & $Q = 2$           & $Q = 4$           & $Q=8$             & $Q=16$             \\
    \midrule 
    Accuracy     & $89.2\pm0.4$      & $89.3\pm0.7$      & $90.7\pm0.5$      & $90.3\pm1.1$      \\
    Runtime       & $1323\pm44$       & $521\pm44$        & $324\pm31$        & $250\pm24$    \\
    \bottomrule
  \end{tabular}
  \vskip -0.1in
\end{table}

\subsection{Flexibility}
To demonstrate the flexibility of \ourmethod, we conduct experiments to show the performance under (a) different GNN backbones and (b) different numbers of clients, $M$.

\begin{figure}[t]
  \centering
  \includegraphics[width = 0.5\linewidth]{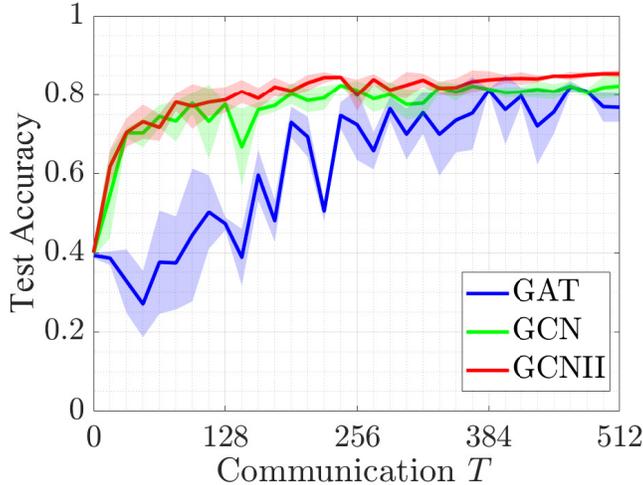}
  \vskip -10pt
  \caption{Test accuracy under three backbone GNNs on PubMed.}
  \label{fig:abl:models}
  \vskip -0.1in
\end{figure}

\textbf{Backbone model:} We compare three backbones: GCN, GAT~\citep{velickovic2018graph}, and GCNII, which are representative GNNs.
The learning rate for each backbone is tuned to its best performance. The test accuracy over training rounds is plotted in \figref{fig:abl:models}. We see that \ourmethod can take different GNNs as the backbone and reach a similar prediction performance, despite that the convergence curves are not all similar. For example, the convergence histories of GCN and GCNII are quite close, whereas that of GAT experiences roughness.

\textbf{Number of clients:} We set $M=3,5,7$ and investigate the change of performance for different training methods. Hyperparameters are tuned to achieve the optimal accuracy under a fixed number of epochs. \tabref{tab:abl:client} suggests that the performance of standalone training decreases as $M$ increases, which is expected because each client has fewer features while server aggregation is not performed. Meanwhile, the performance of \ourmethod is not affected and it stays comparable with that of centralized training. We note that it is unrealistic to set $M$ arbitrarily large, because $M$ is limited by the feature length and also in practice, it is determined by data ownership.

\begin{table}[t]
  \centering
  \caption{Test accuracy (\%) of Centralized training (Cent.), Standalone training (StAl.), and \ourmethod under different numbers of clients ($M =3,5,7$). Top: CiteSeer; bottom: PubMed.}
  \label{tab:abl:client}
  \vskip 5pt
  \small
  \begin{tabular}{l | ccc}
    \toprule
    \# Client   & $M=3$             & $M=5$             & $M=7$             \\
    \midrule
    Cent.       & \multicolumn{3}{c}{\rule[3pt]{.6in}{0.4pt} \,\, $70.2\pm0.8$ \,\, \rule[3pt]{.6in}{0.4pt}}     \\
    StAl.       & $64.4\pm0.5$      & $44.6\pm0.3$      & $36.6\pm 0.8$      \\
    \ourmethod  & $68.8\pm3.3$      & $69.5\pm1.4$      & $69.4\pm0.7$\\
    \bottomrule 
    \toprule
    \# Client   & $M=3$             & $M=5$             & $M=7$             \\
    \midrule
    Cent.       & \multicolumn{3}{c}{\rule[3pt]{.6in}{0.4pt} \,\, $84.9\pm0.6$ \,\, \rule[3pt]{.6in}{0.4pt}}     \\
    StAl.       & $77.2\pm 0.5$      & $73.8\pm0.5$      & $67.0\pm1.3$      \\
    \ourmethod  & $83.8\pm1.8$       & $84.6\pm 0.4$    &$84.4\pm 0.2$   \\
    \bottomrule
  \end{tabular}
  \vskip -0.1in
\end{table}

\section{Conclusion}
We have presented a flexible model splitting approach for VFL with vertically distributed graph data and proposed a communication-efficient algorithm, \ourmethod, to train the resulting GNN. Due to the graph structure, VFL on GNNs incurs heavy communication and poses an extra challenge in the convergence analysis, as the stochastic gradients are no longer unbiased. To overcome these challenges, our approach uses lazy aggregation to skip server-client communication and stale global information to update local models, leading to significant communication reduction. Our analysis makes no assumptions on unbiased gradients. We provide extensive experiments to show the flexibility of the model and the communication saving in training, without compromise on the model quality.

\bibliography{ref}
\bibliographystyle{IEEEbib}

\clearpage
\onecolumn
\appendix
\section{Related Works}\label{app:related}
\textbf{Vertical federated learning} is a learning paradigm where the features of the data are distributed across clients. The model is split among the clients and therefore, the key challenge is the heavy communication burden in exchanging the partial sample information.

\citet{gu2021privacy,chen2020vafl,xu2021fedv,yang2019parallel} consider a linear combination of the features (e.g., SVM, linear regression, logistic regression) and develop asynchronous communication-efficient protocols to aggregate the partial features. These methods can hardly be generalized to complex models in a highly nonlinear nature (e.g., multi-layer perceptrons and convolutional neural networks), because they require multiple rounds of communications. \citet{yang2019quasi} use the second-order Taylor expansion to linearize the training objective for efficient communication. All these methods assume that one round of communication is sufficient for exchanging partial information, too simplistic and ineffective for graph neural networks.

\begin{figure}[h]
  \centering
  \subfigure[Graph-level]{
    \includegraphics[width=0.23\textwidth]{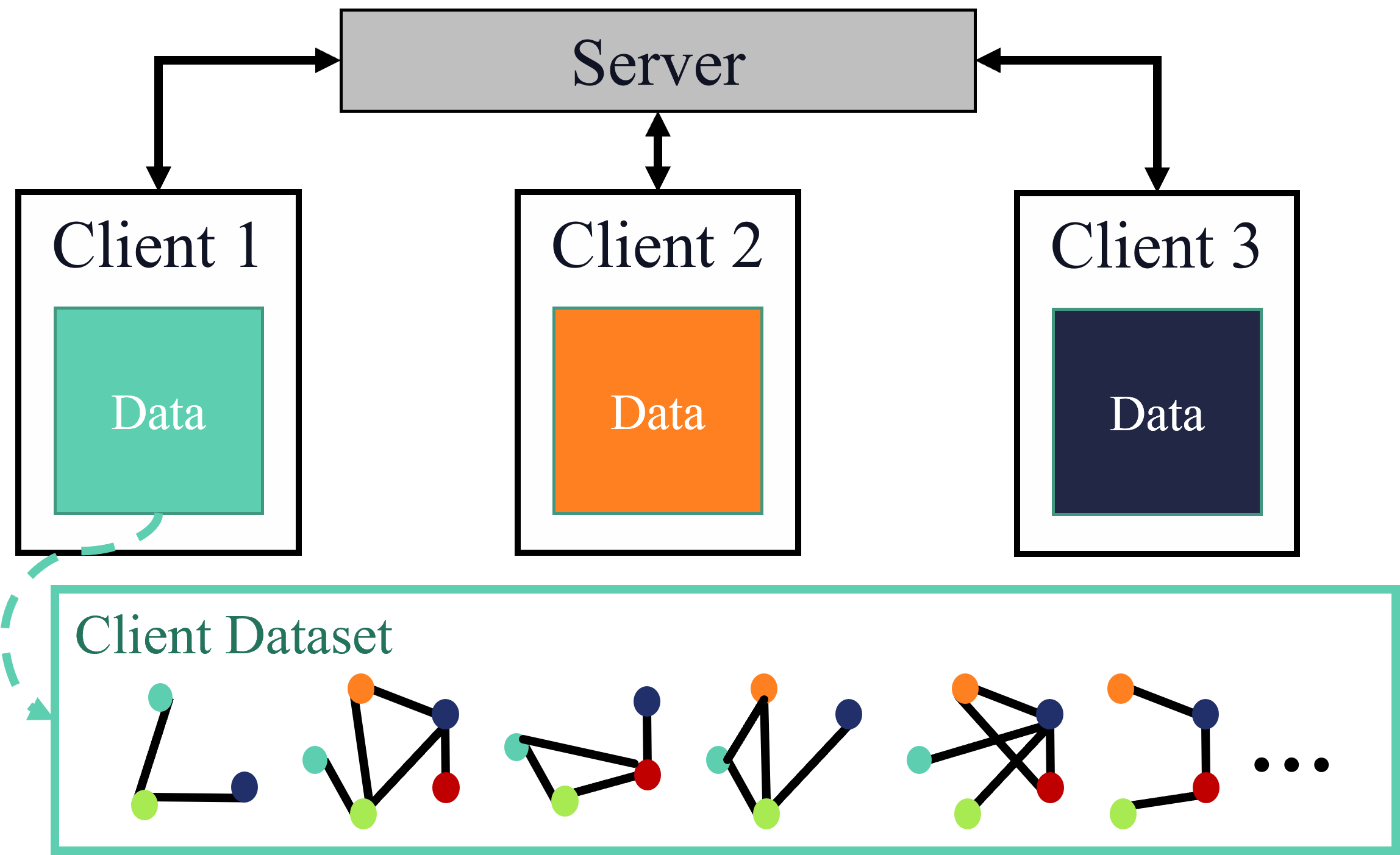}
  }
  \subfigure[Subgraph-level, vertical]{
    \includegraphics[width=0.23\textwidth]{figure/Data_v1.png}
  }
  \subfigure[Subgraph-level, horizontal]{
    \includegraphics[width=0.23\textwidth]{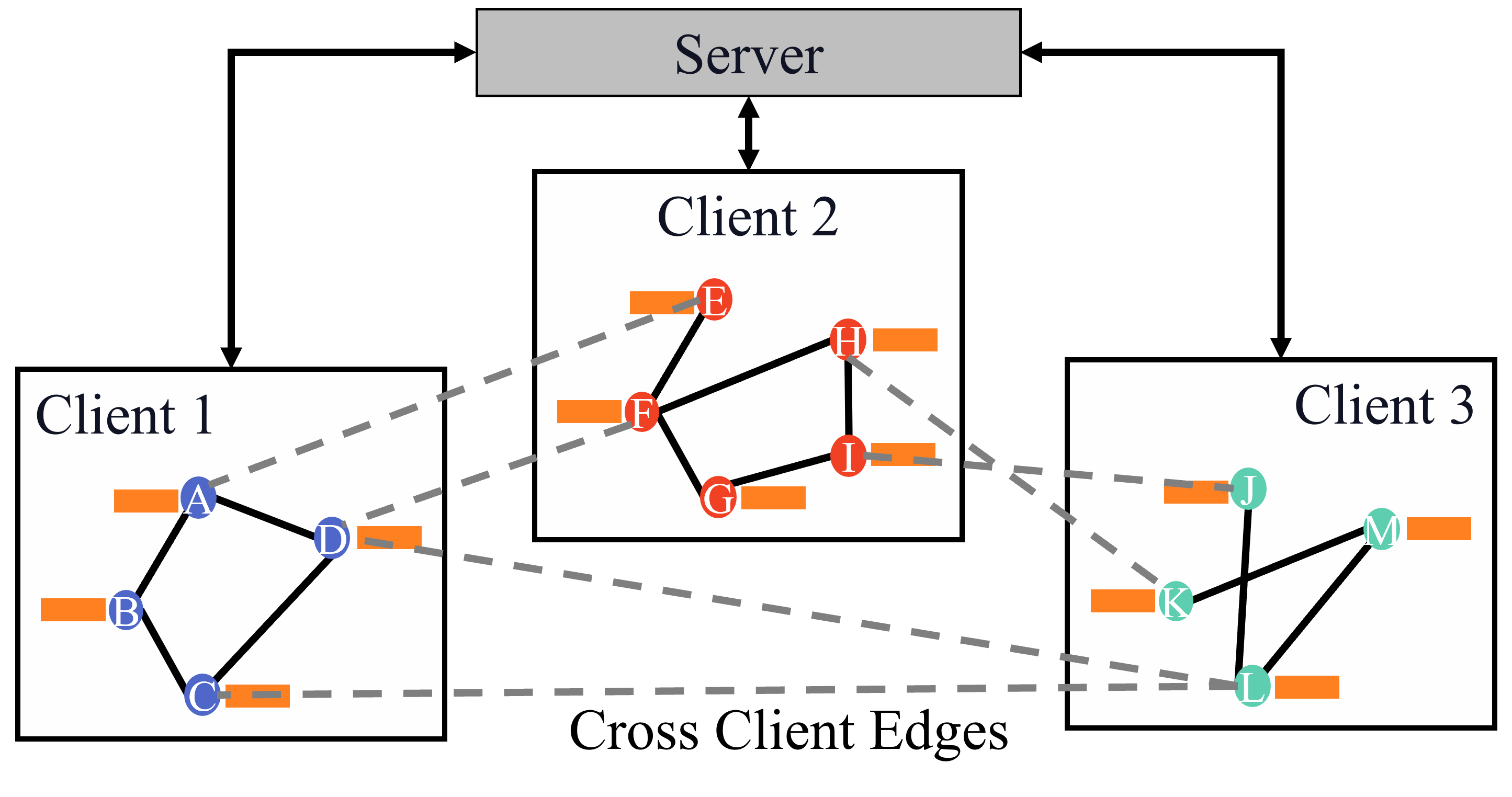}
  }
  \subfigure[Node-level]{
    \includegraphics[width=0.23\textwidth]{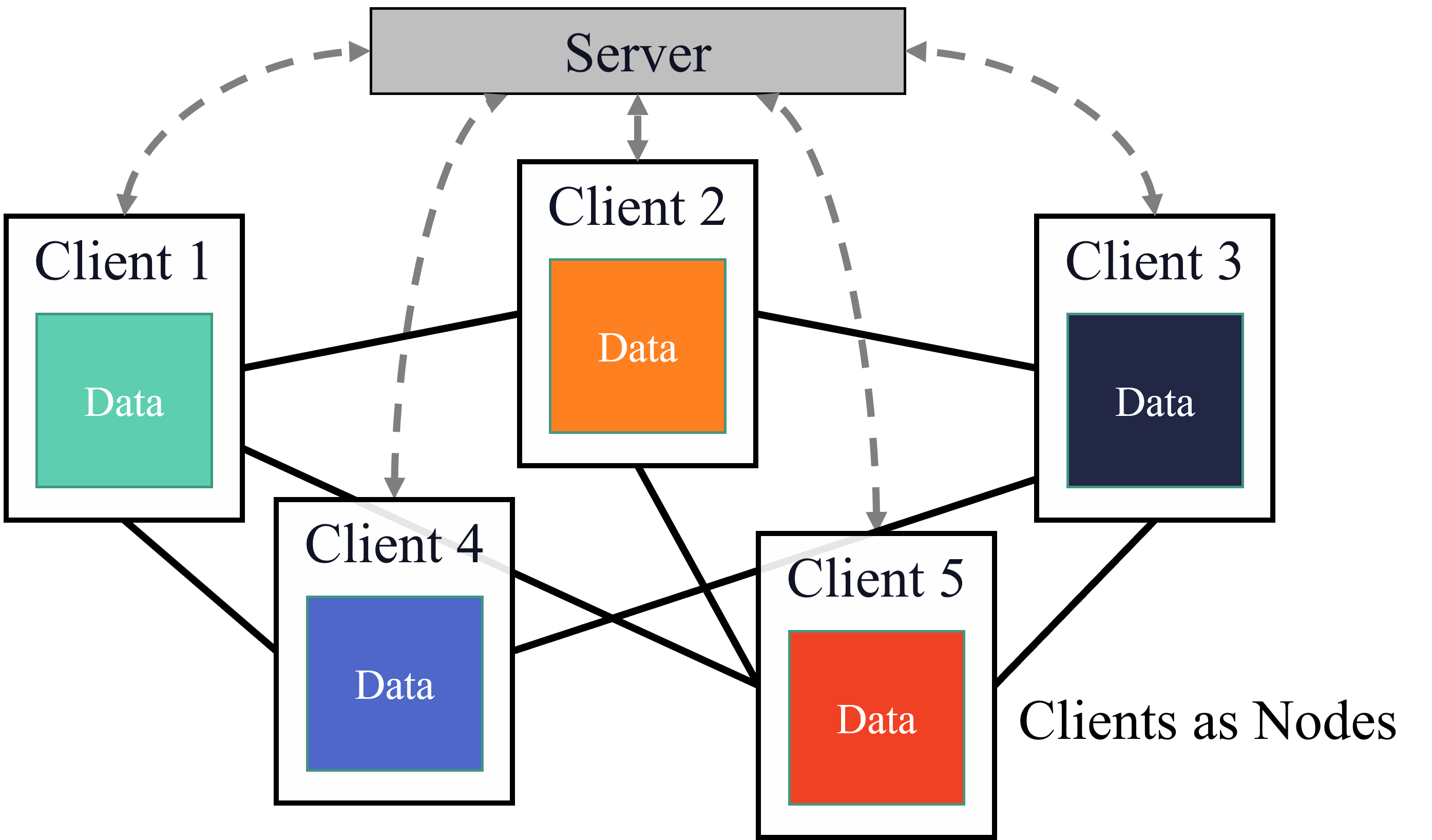}
  }
  \caption{Four scenarios of federated learning with graphs.}
  \label{fig:review}
  \vskip -0.1in
\end{figure}

\textbf{Federated learning with graphs} is a broad subject due to the flexibility of graphs. There are generally four scenarios. Some are related to graph-structured data while some are to graph-structured clients. See \figref{fig:review} for an illustration.

\textbf{(1) Graph-level (horizontal).} In this scenario, each client possesses a number of graphs and all clients collaborate to train a unified model. \citet{xie2021federated} apply federated learning to molecular graph property prediction with decentralized data. \citet{he2021fedgraphnn} design a coding platform for training and evaluating graph-level federated learning algorithms. \citet{bayram2021federated} study the scenario that multiple hospitals collaborate to train a GNN model for connectional brain templates. \citet{wang2020graphfl} use a model-agnostic meta-learning (MAML) approach to deal with heterogeneous graphs and train a common model among clients. \citet{zheng2021asfgnn} study the setting where each client holds a heterogeneous graph dataset, a heterogeneous GNN feature extractor, and a common classifier. In this case, the clients train the local model with local datasets and upload the parameters of the classifier to the server. The server performs model averaging and optimizes the hyperparameters for local training.

Different from the scenario addressed by our paper, each client's data are independent; hence, each client can separately perform local training without gradient sharing.

\textbf{(2) Subgraph-level, vertical.} In this scenario, each client possesses a fraction of the graph information, particularly, partial edges and partial node features. Each client holds a partial model, which is responsible for the features and the induced subgraph it possesses. We focus on this scenario in the paper.

One way for the clients to collaboratively train the model is that they each uses a separate local GNN to extract node representations~\citep{zhou2020vertically}. Then, the server aggregates these representations to make the prediction. In this approach, the local node representations are obtained by using a local graph, failing to capture the information from neighboring nodes in other clients' graphs. Another way of collaboration is to simulate centralized training~\cite{ni2021vertical}. In this approach, the clients communicate the node representations after each layer computation, causing a large communication overhead.


\textbf{(3) Subgraph-level, horizontal.} In this scenario, each client holds a partition of the graph and all features of the nodes in the partition. Because the features are not split across clients, this scenario is not vertical; if we consider each node to be a data sample, it fits the usual horizontal definition of federated learning. A critical challenge to address is how to economically communicate information of neighboring nodes crossing partitions, hop by hop.

In~\citet{chen2021fedgl}, the clients send their local model, the predicted labels, and the node representations to the server for global semi-supervised learning that assists local model training. In~\citet{chen2022fedgraph}, an active node sampling method is proposed along with reinforcement learning, to improve the performance of local stochastic training. In~\citet{yao2022fedgcn}, the clients first communicate the raw aggregated node features necessary for local training and then perform local training with federated model averaging. In~\citet{zhang2021subgraph}, each client holds an ego-graph with missing neighbors and trains a neural network to predict the information of these missing nodes, by federated learning. Then, each client utilizes the generated pseudo neighbor information for GNN training. In~\citet{wu2021fedgnn}, each client holds one user-item ego-graph, part of the entire recommendation system graph. The clients jointly train a unified model with privacy-preserving methods, using differentially private SGD updates and encrypted node sampling.

\textbf{(4) Node-level.} In this scenario, the clients are connected by a (communication) graph. The setting is akin to decentralized learning. The data at hand may be graph-structured or not.

In~\citet{lalitha2019peer,meng2021cross}, the clients run federated learning algorithms locally and communicate models/gradients through the graph. In the latter work, the clients possess interconnected time-series data. Each of them processes the local data and the server utilizes the underlying graph structure to run a GNN to improve global prediction. In~ \cite{caldarola2021cluster,rizk2021graph}, the underlying graph structure is kept by the server, which uses a GNN to perform aggregation on the model or data collected by the clients.

\section{Subroutines}

\subsection{Subroutines in \algref{alg:train}}\label{app:algorithm}
The subroutines referenced in \algref{alg:train} are Algorithms~\ref{alg:sample}--\ref{alg:local_update:2}.

\begin{algorithm}[H]
  \begin{minipage}[t]{0.5\linewidth}
    \begin{algorithmic}[1]
      \STATE {\textbf{Client:}}
      \FOR {$k = K,\dots, 2$}
      \STATE {Receive($\cS[l_k]$)}.
      \STATE Set $\cS_m[l_k] = \cS[l_k]$.
      \FOR {$l = l_k-1,\dots,l_{k-1}$}
      \STATE Uniformly randomly sample indices $\cS_{m}[l]$\\ from neighbors of $\cS_m[l+1]$.
      \ENDFOR
      \STATE {Send($\cS_m[l_{k-1}]$)} {\textbf{if} $k>2$}.
      \ENDFOR
      \STATE \textbf{Output:} $\{\cS_m[l]\}^L_{l=0}$
    \end{algorithmic}
  \end{minipage}
  \begin{minipage}[t]{0.5\linewidth}
    \begin{algorithmic}[1]
      \STATE {\textbf{Server:}}
      \STATE Uniformly and independently sample \\ \quad indices $\cS[L]$ from training set.
      \STATE Broadcast($\cS[L]$).
      \FOR {$k = K-1,\dots, 2$}
      \STATE Aggregate($\cS_{m}[l_{k}]$).
      \STATE Compute $\cS[l_{k}] = \bigcup^M_{m=1}\cS_{m}[l_{k}]$.
      \STATE Broadcast($\cS[l_{k}]$).
      \ENDFOR
    \end{algorithmic}
  \end{minipage}
  \caption{Sampling Procedure}
  \label{alg:sample}
\end{algorithm}

\begin{algorithm}[H]
  \begin{minipage}[t]{0.55\linewidth}
    \begin{algorithmic}[1]
      \STATE {\textbf{Client:}}
      \STATE \textbf{Input:} $\mW_m, \cD_m, \{\cS_m[l]\}^L_{l=0}$
      \STATE Set $\mH_m[0] = \mX_m[\cS_m[0]]$
      \FOR {$l = 0,\dots, L-1$}
      \STATE $\mH^+_m[l] = \sigma(\mA(\cE(\cS_m[l+1], \cS_m[l]))\mH_m[l]\mW_m[l])$
      \IF{$l \in \cI$}
      \STATE {Send $\mH^+_m[l]$ to server}
      \STATE {Receive $\mH_m[l+1]$}
      \STATE $\mH_{-m}[l+1] = \mathrm{Extract}(\mH_m[l+1], \mH^+_m[l])$
      \ELSE
      \STATE {Set $\mH_m[l+1] = \mH_m^+[l]$}
      \ENDIF
      \ENDFOR
      \STATE \textbf{Output:} $\{\mH_{-m}[l+1]\}_{l\in\cI}$
    \end{algorithmic}
  \end{minipage}
  \begin{minipage}[t]{0.45\linewidth}
    \begin{algorithmic}[1]
      \STATE {\textbf{Server:}}
      \FOR {$l \in \cI$}
      \STATE $\mH[l+1] =\mathrm{Agg}(\mH_1^+[l],\dots, \mH_M^+[l])$.
      \STATE Broadcast $\mH[l+1]$.
      \ENDFOR
    \end{algorithmic}
  \end{minipage}
  \caption{JointInference}
  \label{alg:local_update:1}
\end{algorithm}

\begin{algorithm}[H]
  \begin{algorithmic}[1]
    \STATE \textbf{Input:} $\mW_m^{t,q}, \cD_m, \{\cS_m^t[l]\}^L_{l=0}, \{\mH^t_{-m}[l+1]\}_{l\in\cI}$
    \STATE Set $\mH^{t,q}_m[0] = \mX_m[\cS^t_m[0]]$
    \FOR {$l = 0,\dots, L-1$}
    \STATE $\mH^{t,q,+}_m[l] = \sigma(\mA(\cE(\cS^{t}_m[l+1], \cS^{t}_m[l]))\mH^{t,q}_m[l]\mW^{t,q}_m[l])$
    \IF{$l \in \cI$}
    \STATE {Set $\mH^{t,q}_m[l+1] = \mathrm{Agg}(\mH_{-m}^{t}[l+1], \mH_m^{t,q,+}[l])$}
    \ELSE
    \STATE {Set $\mH_m^{t,q}[l+1] = \mH_m^{t,q,+}[l]$}
    \ENDIF
    \ENDFOR
    \STATE Compute loss $\cL_m^{t,q} = \ell\left(\yy[\cS^t_m[L]], f_m(\mH^{t,q}_m[L], \mW_m[L])\right)$
    \STATE {\bf Output:} $\mW^{t,q+1}_m = \mW_m^{t,q} - \eta^{t,q}\nabla_{\mW_m^{t,q}}\cL_m^{t,q}$
  \end{algorithmic}
  \caption{LocalUpdate}
  \label{alg:local_update:2}
\end{algorithm}

\subsection{Modified Subroutines When Only One Client Holds the Labels}\label{app:algorithm_v2}
Without loss of generality, we assume that the labels are held by client $m=1$. Algorithms~\ref{alg:sample}--\ref{alg:local_update:2} are modified to Algorithms~\ref{alg:sample_v2}--\ref{alg:local_update:2_v2}, respectively.

For sampling (\algref{alg:sample_v2}), client 1, rather than the server, generates the sampled node indices for training.

For JointInference (\algref{alg:local_update:1_v2}), client 1 computes the mini-batch loss and the partial gradient with respect to the node representation of the last layer, $\mH_1[L]$, and broadcasts it to all other clients through the server. We require that server aggregation is performed on the last layer; that is, for other nodes, $\mH_m[L] = \mathrm{Agg}(\mH^+_1[L-1],\dots, \mH^+_{M}[L-1])$. Then, the partial gradient with respect to $\mH_1[L]$ (that is, $\nabla_{\mH_1[L]}\cL_1$) is sufficient for computing the gradients of the model parameters on all these clients. To see this, note that the clients $m>1$ do not have the classifier $\mW_m[L]$. Hence, we have
\begin{align}
    \nabla_{\mW_m}\cL_m &= \frac{\partial}{\partial \mW_m} \ell\left(\yy[\cS_1[L]], f_1(\mH_m[L], \mW_1[L])\right)\nonumber\\
    & = \lin{\frac{\partial}{\partial \mH_m[L]} \ell\left(\yy[\cS_1[L]], f_1(\mH_m[L], \mW_1[L])\right), \frac{\partial}{\partial \mW_m}\mH_m[L]}\nonumber\\
    & = \lin{\frac{\partial\cL_1}{\partial \mH_1[L]}, \frac{\partial \mH_1[L]}{\partial \mW_m[L]}}.\label{eq:backprop}
\end{align}
Therefore, client 1 broadcasts $\nabla_{\mH_1[L]}\cL_1$ to all other clients through the server.

For LocalUpdate (\algref{alg:local_update:2_v2}), clients $m\geq 2$ follow the chain rule to compute the second term inside the inner product of~\eqref{eq:backprop} and update the local model parameters.

\begin{algorithm}[H]
  \begin{minipage}[t]{0.5\linewidth}
    \begin{algorithmic}[1]
      \STATE {\textbf{Client:}}
      \IF{$m=1$}
      \STATE Uniformly and independently sample \\ \quad indices $\cS[L]$ from training set.
      \ENDIF
      \FOR {$k = K,\dots, 2$}
      \STATE {Receive($\cS[l_k]$)}.
      \STATE Set $\cS_m[l_k] = \cS[l_k]$.
      \FOR {$l = l_k-1,\dots,l_{k-1}$}
      \STATE Uniformly randomly sample indices $\cS_{m}[l]$\\ from neighbors of $\cS_m[l+1]$.
      \ENDFOR
      \STATE {Send($\cS_m[l_{k-1}]$)} {\textbf{if} $k>2$}.
      \ENDFOR
      \STATE \textbf{Output:} $\{\cS_m[l]\}^L_{l=0}$
    \end{algorithmic}
  \end{minipage}
  \begin{minipage}[t]{0.5\linewidth}
    \begin{algorithmic}[1]
      \STATE {\textbf{Server:}}
      \STATE Recieve sample indices $\cS[L]$ from client $1$.
      \STATE Broadcast($\cS[L]$).
      \FOR {$k = K-1,\dots, 2$}
      \STATE Aggregate($\cS_{m}[l_{k}]$).
      \STATE Compute $\cS[l_{k}] = \bigcup^M_{m=1}\cS_{m}[l_{k}]$.
      \STATE Broadcast($\cS[l_{k}]$).
      \ENDFOR
    \end{algorithmic}
  \end{minipage}
  \caption{Sampling Procedure (Modified)}
  \label{alg:sample_v2}
\end{algorithm}

\begin{algorithm}[H]
  \begin{minipage}[t]{0.55\linewidth}
    \begin{algorithmic}[1]
      \STATE {\textbf{Client:}}
      \STATE \textbf{Input:} $\mW_m, \cD_m, \{\cS_m[l]\}^L_{l=0}$
      \STATE Set $\mH_m[0] = \mX_m[\cS_m[0]]$
      \FOR {$l = 0,\dots, L-1$}
      \STATE $\mH^+_m[l] = \sigma(\mA(\cE(\cS_m[l+1], \cS_m[l]))\mH_m[l]\mW_m[l])$
      \IF{$l \in \cI$}
      \STATE {Send $\mH^+_m[l]$ to server}
      \STATE {Receive $\mH_m[l+1]$}
      \STATE $\mH_{-m}[l+1] = \mathrm{Extract}(\mH_m[l+1], \mH^+_m[l])$
      \ELSE
      \STATE {Set $\mH_m[l+1] = \mH_m^+[l]$}
      \ENDIF
      \ENDFOR
      \IF{$m=1$}
      \STATE Compute loss $\cL_1^{t,0} = \ell\left(\yy[\cS^t_1[L]], f_1(\mH^{t,0}_1[L], \mW_1[L])\right)$
      \STATE Send $\nabla_{\mH_1^{t,0}[L]}\cL_1^{t,0}$ to server
      \ELSE
      \STATE Receive $\nabla_{\mH_1^{t,0}[L]}\cL_1^{t,0}$ from server
      \ENDIF
      \STATE \textbf{Output:} $\{\mH_{-m}[l+1]\}_{l\in\cI}, \nabla_{\mH_1^{t,0}[L]}\cL_1^{t,0}$
    \end{algorithmic}
  \end{minipage}
  \begin{minipage}[t]{0.45\linewidth}
    \begin{algorithmic}[1]
      \STATE {\textbf{Server:}}
      \FOR {$l \in \cI$}
      \STATE $\mH[l+1] =\mathrm{Agg}(\mH_1^+[l],\dots, \mH_M^+[l])$.
      \STATE Broadcast $\mH[l+1]$.
      \ENDFOR
      \STATE Receive $\nabla_{\mH_1^{t,0}[L]}\cL_1^{t,0}$ from client $1$
      \STATE Broadcast $\nabla_{\mH_1^{t,0}[L]}\cL_1^{t,0}$.
    \end{algorithmic}
  \end{minipage}
  \caption{JointInference (Modified)}
  \label{alg:local_update:1_v2}
\end{algorithm}

\begin{algorithm}[H]
  \begin{algorithmic}[1]
    \STATE \textbf{Input:} $\mW_m^{t,q}, \cD_m, \{\cS_m^t[l]\}^L_{l=0}, \{\mH^t_{-m}[l+1]\}_{l\in\cI}, \nabla_{\mH_1^{t,0}[L]}\cL_1^{t,0}$
    \STATE Set $\mH^{t,q}_m[0] = \mX_m[\cS^t_m[0]]$
    \FOR {$l = 0,\dots, L-1$}
    \STATE $\mH^{t,q,+}_m[l] = \sigma(\mA(\cE(\cS^{t}_m[l+1], \cS^{t}_m[l]))\mH^{t,q}_m[l]\mW^{t,q}_m[l])$
    \IF{$l \in \cI$}
    \STATE {Set $\mH^{t,q}_m[l+1] = \mathrm{Agg}(\mH_{-m}^{t}[l+1], \mH_m^{t,q,+}[l])$}
    \ELSE
    \STATE {Set $\mH_m^{t,q}[l+1] = \mH_m^{t,q,+}[l]$}
    \ENDIF
    \ENDFOR
    \IF{$m=1$}
    \STATE Compute loss $\cL_1^{t,q} = \ell\left(\yy[\cS^t_m[L]], f_m(\mH^{t,q}_m[L], \mW_m[L])\right)$
    \ELSE
    \STATE Compute partial loss based on $\nabla_{\mH_1^{t,0}[L]}\cL_1^{t,0}$ by using the chain rule: $\cL_m^{t,q} = \nabla_{\mH^{t,q}_m[L]}{\cL^{t,0}_1}$
    \ENDIF
    \STATE {\bf Output:} $\mW^{t,q+1}_m = \mW_m^{t,q} - \eta^{t,q}\nabla_{\mW_m^{t,q}}\cL_m^{t,q}$
  \end{algorithmic}
  \caption{LocalUpdate (Modified)}
  \label{alg:local_update:2_v2}
\end{algorithm}

\section{Proofs for Section \ref{sec:theory:converge}}\label{app:theory}

\subsection{Assumptions}\label{app:theory:as}

\begin{assumption}[Smooth function and Lipschitz gradient]\label{as:smooth}
    The loss function $\ell$ is $G_\ell$-smooth with $L_\ell$-Lipschitz gradient, i.e.,
    \begin{align*}
        \norm{\ell(\yy, \cS, \mW) - \ell(\yy, \cS, \mW')} &\leq G_\ell\norm{\mW - \mW'}\\
        \norm{\nabla_\mW\ell(\yy, \cS, \mW) - \nabla_{\mW'}\ell(\yy, \cS, \mW')} &\leq L_\ell\norm{\mW - \mW'}, \quad \forall~\mW,\mW'
    \end{align*}
    and each client's prediction function $f_m$ is $G_f$-smooth with $L_f$-Lipschitz gradient, i.e.,
    \begin{align*}
        \norm{f_m(\cS, \mW_m) - f_m(\cS, \mW'_m)} &\leq G_f\norm{\mW_m - \mW'_m}\\
        \norm{\nabla_{\mW_m} f_m(\cS, \mW_m) - \nabla_{\mW'_m} f_m(\cS, \mW_m)} &\leq L_f\norm{\mW_m - \mW'_m}, \quad \forall~\mW_m, \mW'_m, \forall~m.
    \end{align*}
\end{assumption}

\begin{assumption}[Lower-bounded objective]\label{as:lower_bound}
    The training objective is bounded below; that is, there exists a constant $\cL^\star > -\infty$ such that for all $\{\mW_m\}$, it satisfies that \[\cL(\{\mW_m\}) \geq \cL^\star.\]
\end{assumption}

\begin{assumption}[Uniform sampling]\label{as:sample}
    At each iteration $t$,
    the server and the clients uniformly sample nodes $\{\cS_m[l]\}^L_{l=0}$, with $|\cS[L]|=S$, according to \algref{alg:sample}.
\end{assumption}

\subsection{Proof of Theorem~\ref{thm:convergence}}\label{app:theory:converge}

We first note the following useful relation:
\begin{equation}\label{eq:ab}
    \begin{aligned}
        &\norm{a+b}^2=\norm{a-c + c-b}^2 \leq (1+\alpha)\norm{a-c}^2 + (1+\frac{1}{\alpha})\norm{c-b}^2, \quad\forall \alpha>0.
    \end{aligned}
\end{equation}
For notation simplicity, let us denote the expectation conditioned on all the information before iteration $t$ as \[\E^t[~\cdot~] = \E_{\cS^t}[~\cdot~| \mW^{t-1,Q},\dots, \mW^{0,0}, \cS^{t-1}, \dots, \cS^0];\] denote the ``all-but-$m$'' vector as $(\cdot)_{-m},$ (e.g., the collection of all client parameters except for client $m$ is $\mW_{-m} = \{\mW_{m'}\}_{m'\neq m}$); denote the client model updated with data $\cS$ as $\mW_m(\cS)$; denote the gradient evaluated with data $\cS$ on parameter $\mW_m$ as $\nabla\cL(\mW_m(\cS),\cS)$; and denote the stacked gradient of all clients as $\mG = [\nabla\cL(\mW_1(\cS),\cS), \dots, \nabla\cL(\mW_M(\cS),\cS)].$ Then, the update rule can be rewritten as:
\begin{equation}\label{eq:update_S}
    \mW^{t,q+1}(\cS^t) = \mW^{t,q}(\cS^t) - \eta \mG^{t,q}.
\end{equation} In addition, let us define a virtual model sequence updated with {\it full data} as $\mW(\cD)$, i.e., 
\begin{equation}\label{eq:update_D}
    \mW^{t,q+1}(\cD) = \mW^{t,q}(\cD) - \eta \nabla\cL(\mW^{t,q}(\cD), \cD).
\end{equation}

We can bound the variance of the stochastic gradient at any round $t$ and iteration $q=0$ with the following lemma:
\begin{lemma}[Bounded variance]\label{le:variance}
  Under Assumptions~\ref{as:smooth}--\ref{as:sample}, with probability at least $p = 1-\delta$, the variance of the stochastic gradient is bounded by:
  \begin{equation}
    \E^t\left[\norm{\nabla \cL(\mW; \cS^t) - \nabla\cL(\mW; \cD)}^2\right] \leq \sigma, \quad \forall~\mW \text{ independent of }\cS^t,
  \end{equation}
  where
  \begin{equation}
    \sigma = 64G_\ell^2L_f^2\log\left(\frac{2d}{\delta}\right) + 128L_\ell^2\left(G_f^4+ \frac{1}{S}\right)\left(\log\left(\frac{2d}{\delta}\right)+ \frac{1}{4}\right).
  \end{equation}
\end{lemma}
The main technique for proving this lemma is to use the matrix Bernstein inequality~\citep{tropp2015introduction} to bound the variance of the stochastic gradients and the variance of the expectation for each client. The proof steps of \leref{le:variance} follows the same steps in the proofs for Lemmas 5 and 6 of \citet{ramezani2020gcn}, so we omit them here.

Further, we bound the Lipschitz constant of the total loss function in the following lemma:
\begin{lemma}[Lipschitz gradient]\label{le:smooth}
  Under Assumptions~\ref{as:smooth}--\ref{as:sample}, the full gradient and each partial gradient of the objective $\cL(\mW, \cS)$ are Lipschitz continuous with uniform constant $C_0 = G_\ell L_f + G_f^2L_\ell$:
  \begin{align*}
      \norm{\nabla_{\mW}\cL(\mW, \cS) - \nabla_{\mW'}\cL(\mW', \cS)} &\leq C_0 \norm{\mW - \mW'}, \quad\forall~\mW, \mW'\\
      \norm{\nabla_{\mW_m}\cL(\mW, \cS) - \nabla_{\mW'_m}\cL(\mW', \cS)} &\leq C_0 \norm{\mW - \mW'}, \quad\forall~\mW, \mW', \forall~m.
  \end{align*}
\end{lemma}

The proof of \leref{le:smooth} is given below in \secref{app:proof:le}.

With the above results, we begin our proof for Theorem \ref{thm:convergence}. First, applying \leref{le:smooth}, we have:
\begin{equation}\small
    \begin{aligned}
        & \cL(\mW^{t,q+1}, \cD) - \cL(\mW^{t,q}, \cD)\leq \lin{\nabla\cL(\mW^{t,q}, \cD), \mW^{t,q+1}-\mW^{t,q}}+\frac{C_0}{2}\norm{\mW^{t,q+1}-\mW^{t,q}}^2\\
        & \stackrel{(a)}{=}  -\eta\lin{\nabla\cL(\mW^{t,q}, \cD), \mG^{t,q}}+\frac{C_0\eta^2}{2}\norm{\mG^{t,q}}^2\\
        & \stackrel{(b)}{=} -\frac{\eta}{2}\left(\norm{\nabla\cL(\mW^{t,q}, \cD)}^2+\norm{\mG^{t,q}}^2 - \norm{\nabla \cL(\mW^{t,q} ,\cD)-\mG^{t,q}}^2\right) +\frac{C_0\eta^2}{2}\norm{\mG^{t,q}}^2\\
        & =  -\frac{\eta}{2}\norm{\nabla\cL(\mW^{t,q}, \cD)}^2-\frac{\eta}{2}(1-\eta C_0)\norm{\mG^{t,q}}^2 + \frac{\eta}{2}\norm{\nabla \cL(\mW^{t,q}, \cD)-\mG^{t,q}}^2,
    \end{aligned}
\end{equation}
where step $(a)$ applies the update rule of \algref{alg:local_update:2} and step $(b)$ uses the fact that $\lin{a,b} = \frac{1}{2}\left(\norm{a}^2+\norm{b}^2 - \norm{a-b}^2\right)$. Taking expectation, we have:
\allowdisplaybreaks
    \begin{align}
        &\E^{t}[\cL(\mW^{t,q+1}, \cD)- \cL(\mW^{t,q}, \cD)] \leq  -\frac{\eta}{2}\E^t\norm{\nabla\cL(\mW^{t,q}, \cD)}^2\nonumber\\
        &\quad -\frac{\eta}{2}(1-\eta C_0)\E^t\norm{\mG^{t,q}}^2 + \frac{\eta}{2}\E^t\norm{\nabla \cL(\mW^{t,q}, \cD)-\mG^{t,q}}^2\nonumber\\
        & \stackrel{(a)}{=} -\frac{\eta}{2}\E^t\norm{\nabla\cL(\mW^{t,q}, \cD)}^2 + \frac{\eta}{2}\E^t\norm{\nabla \cL(\mW^{t,q}, \cD)-\mG^{t,q}}^2  \nonumber\\
        &\quad -\frac{\eta}{2}(1-\eta C_0)(\norm{\E^t\mG^{t,q}}^2+\E^t\norm{\mG^{t,q}-\E^t\mG^{t,q}}^2)\nonumber\\
        & \stackrel{(b)}{\leq}-\frac{\eta}{2}\E^t\norm{\nabla\cL(\mW^{t,q},\cD)}^2 -\frac{\eta}{2}(1-\eta C_0)(\norm{\E^t\mG^{t,q}}^2+\E^t\norm{\mG^{t,q}-\E^t\mG^{t,q}}^2)\nonumber\\
        & \quad + \frac{\eta}{2}\left((1+\frac{1}{\eta C_0})\E^t\norm{\nabla \cL(\mW^{t,q}, \cD)-\E^t\mG^{t,q}}^2+ (1+\eta C_0)\E^t\norm{\E^t\mG^{t,q} - \mG^{t,q}}^2\right)\nonumber\\
        & = -\frac{\eta}{2}\E^t\norm{\nabla\cL(\mW^{t,q}, \cD)}^2 -\frac{\eta}{2}(1-\eta C_0)\norm{\E^t\mG^{t,q}}^2 + \eta^2C_0\underbrace{\E^t\norm{\mG^{t,q}-\E^t\mG^{t,q}}^2}_{\text{Term 1}}\nonumber\\
        & \quad + \frac{1+\eta C_0}{2C_0}\underbrace{\E^t\norm{\nabla \cL(\mW^{t,q}, \cD)-\E^t\mG^{t,q}}^2}_{\text{Term 2}},\label{eq:descent}
    \end{align}
where step $(a)$ uses the fact that $\E(X)^2 = \E(X^2) + \E(X-\E(X))^2$ and step $(b)$ uses \eqref{eq:ab} with $\alpha = \eta C_0$. Next, we bound Term 1 and Term 2 in the above inequality separately.

\subsubsection{Bound of Term 1}\label{app:analysis:part1}
First, \comment{ we denote the variables updated using mini-batch $\cS_l$ as $\mW^{t,q}(\cS_l)$, using sample $\xi$ as $\mW^{t,q}(\xi)$ starting from $\mW^{t,0}$.  That is, we have the following update rules: 
\begin{align}
    \mW^{r_0+1}_{k}(\cS_l)&\bydef\mW^{t,0}_k-\eta g_k(\mW^{t,0};\cS_l),\label{eq:update_sl}\\
    \mW^{t,q}_{k}(\cS_l)&\bydef\mW^{r_0+\tau-1}_{k}(\cS_l)-\eta g_k(\yy^{r_0+\tau-1}_{k}(\cS_l);\cS_l),\nonumber
\end{align}
where $\yy^{r}_{k}(\cS_l)\bydef{[\mW^{t,0}_{-k},\theta^{t,q}_{k}(\cS_l)]}$ is the model used for updating the parameters of party $k$, and
\begin{align}
    \mW^{r_0+1}_{k}(\xi)&\bydef\mW^{t,0}_k-\eta g_k(\mW^{t,0};\xi),\label{eq:update_xi}\\
    \mW^{t,q}_{k}(\xi)&\bydef\mW^{r_0+\tau-1}_{k}(\xi)-\eta g_k(\yy^{r_0+\tau-1}_{k}(\xi);\xi),\nonumber
\end{align}
where $\yy^{r}_{k}(\xi)\bydef{[\mW^{t,0}_{-k},\theta^{t,q}_{k}(\xi)]}$. Additionally, we have $\yy^{t,0}_k = \mW^{t,0}, \forall k\in[K].$
Further let us define $r \triangleq r_0+\tau$.

Using the above notations,} we can rewrite $\E^t[\norm{\mG^{t,q}-\E^t\mG^{t,q}}^2]$ as:
\begin{equation}
    \begin{aligned}
        &\E^t[\norm{\mG^{t,q}-\E^t\mG^{t,q}}^2] = \sum^M_{m=1}\E^t\left[\norm{\nabla \cL(\mW_m^{t,q}(\cS^t), \cS^t) - \E_{\cS}\nabla \cL(\mW_m^{t,q}(\cS), \cS)}^2\right]\\
        &\stackrel{(a)}{\leq}\sum^M_{m=1}\underbrace{\E^t\left[\norm{\nabla \cL(\mW_m^{t,q}(\cS^t), \cS^t) - \nabla \cL(\mW_m^{t,q}(\cD), \cD)}^2\right]}_{\bydef A^{t,q}_m},
    \end{aligned}
\end{equation}
where step $(a)$ uses the fact that $\E(X-\E(X))^2 \leq \E(X - Y)^2$ for all constant $Y$. Then, we can bound $A^{t,q}_m$ as follows. When $q = 0$, by \leref{le:variance}, we obtain that $A^{t,0}_m \leq \sigma$ holds with probability $1-\delta.$ In general, when $q\geq 1,$ we have:
\begin{equation}\small
    \begin{aligned}
        A^{t,q}_m &\stackrel{\eqref{eq:ab}}{\leq} 2\E^t\left[\norm{\nabla \cL(\mW_m^{t,q}(\cS^t), \cS^t) - \nabla \cL(\mW_m^{t,q}(\cD), \cS^t)}^2\right] \\
        &\quad + 2\E^t\left[\norm{\nabla \cL(\mW_m^{t,q}(\cD), \cS^t) - \nabla \cL(\mW_m^{t,q}(\cD), \cD)}^2\right]\\
        & \stackrel{(a)}{\leq} 2C_0^2\E^t\left[\norm{\mW_m^{t,q}(\cS^t) -\mW_m^{t,q}(\cD)}^2\right] + 2\E^t\left[\norm{\nabla \cL(\mW_m^{t,q}(\cD), \cS^t) - \nabla \cL(\mW_m^{t,q}(\cD), \cD)}^2\right]\\
        & \stackrel{(b)}{\leq} 2C_0^2\E^t\left[\norm{\mW_m^{t,q}(\cS^t) -\mW_m^{t,q}(\cD)}^2\right] + 2\sigma,
    \end{aligned}
\end{equation}
which holds with probability $1-\delta$. Here, step $(a)$ applies \leref{le:smooth} to the first term and step $(b)$ applies \leref{le:variance} to the second term. Then, we bound $\E^t\left[\norm{\mW_m^{t,q}(\cS^t) -\mW_m^{t,q}(\cD)}^2\right]$ in the above equation as:
{\allowdisplaybreaks
    \begin{align}
        & \E^t\left[\norm{\mW_m^{t,q}(\cS^t) -\mW_m^{t,q}(\cD)}^2\right] \nonumber\\
        &\stackrel{(a)}{=} \E^t\left[\norm{\mW_m^{t,0} - \eta\sum^{q-1}_{q'=0}\nabla \cL(\mW_m^{t,q'}(\cS^t), \cS^t) - \left(\mW_m^{t,0} - \eta\sum^{q-1}_{q'=0}\nabla \cL(\mW_m^{t,q'}(\cD), \cD)\right)}^2\right]\nonumber\\
        & \stackrel{(b)}{=} \eta^2\E^t\left[\norm{\sum^{q-1}_{q'=0}\left(\nabla \cL(\mW_m^{t,q'}(\cS^t), \cS^t)-\nabla \cL(\mW_m^{t,q'}(\cD), \cD)\right)}^2\right] \nonumber\\
        & \stackrel{(c)}{\leq} \eta^2Q\sum^{q-1}_{q'=0}\E^t\left[\norm{\nabla \cL(\mW_m^{t,q'}(\cS^t), \cS^t)-\nabla \cL(\mW_m^{t,q'}(\cD), \cD)}^2\right] \nonumber\\
        & = \eta^2q\sum^{q-1}_{q'=0}A^{t,q'}_m,
    \end{align}
}where in step $(a)$ we expand the updates to $\mW^{t,0}_m$ with \eqref{eq:update_S} and \eqref{eq:update_D}; step $(b)$ cancels $\mW^{t,0}_m$ and rearrange the terms; and step $(c)$ applies the Cauchy–Schwarz inequality.
At this point, we have the following relations:
\begingroup
\allowdisplaybreaks
    \begin{align*}
        \E^t[\norm{\mG^{t,q}-\E^t\mG^{t,q}}^2] & \leq \sum^M_{m=1} A^{t,q}_m, \quad A^{t,0}_0 \leq \sigma, \quad A^{t,q}_m \leq 2C_0^2\eta^2q\sum^{q-1}_{q'=0}A^{t,q'}_m + 2\sigma, \forall~q\geq 1.
    \end{align*}
\endgroup
Note that $q \leq Q$. By choosing $2\eta^2C_0^2Q^2 \leq 1$, which implies that $\eta \leq \frac{1}{\sqrt{2}QC_0}$, and by recursively substituting the terms, we have the following bounds:
\begin{equation}\label{eq:term1}
    \begin{aligned}
        &A^{t,q}_m \leq \left[2+4q^2\eta^2C_0^2+\frac{8}{3}q^3\eta^4C_0^4\right]\cdot \sigma \leq \frac{14}{3}\sigma,\\
        &\E^t[\norm{\mG^{t,q}-\E^t\mG^{t,q}}^2 \leq M\cdot\left[2+4q^2\eta^2C_0^2+\frac{8}{3}q^3\eta^4C_0^4\right]\cdot \sigma \leq \frac{14M\sigma}{3}.
    \end{aligned}
\end{equation}
This completes bounding the term $\E[\norm{\mG^{t,q}-\E^t\mG^{t,q}}^2]$.

\subsubsection{Bound of Term 2}\label{app:analysis:part2}
We have the following series of relations:
{\allowdisplaybreaks
    \begin{align}
        &\E^t\norm{\nabla \cL(\mW^{t,q}, \cD)-\E^t\mG^{t,q}}^2 = \sum^M_{m=1}\E^t\norm{\nabla_{\mW_m}\cL(\mW^{t,q}(\cS^t), \cD) - \E_{\cS}\nabla \cL(\mW^{t,q}_m(\cS), \cS)}^2\nonumber\\
        & \stackrel{(a)}{\leq} \sum^M_{m=1}\E^t\E_{\cS}\norm{\nabla_{\mW_m}\cL(\mW^{t,q}(\cS^t), \cS) - \E_{\cS}\nabla \cL(\mW^{t,q}_m(\cS), \cS)}^2\nonumber\\
        & \stackrel{(b)}{\leq} \sum^M_{m=1}C_0^2\E^t\E_{\cS}\norm{\mW^{t,q}(\cS^t) - [\mW^{t,q}_m(\cS),\mW^{t,0}_{-m}]}^2\nonumber\\
        & = \sum^M_{m=1}C_0^2\E^t\E_{\cS}\left[\norm{\mW^{t,q}_m(\cS^t) - \mW^{t,q}_m(\cS)}^2 + \sum_{m'\neq m}\norm{\mW^{t,q}_{m'}(\cS^t) - \mW^{t,0}_{m'}}^2\right]\nonumber\\
        & \stackrel{(c)}{=} \eta^2\sum^M_{m=1}C_0^2\E^t\E_{\cS}\left[\norm{\sum^{q-1}_{q' =0}\left(\nabla \cL(\mW^{t,q'}_m(\cS^t),\cS^t) - \nabla \cL(\mW^{t,q'}_m(\cS),\cS)\right)}^2\right.\nonumber\\
        &\qquad \left.+ \sum_{m'\neq m}\norm{\sum^{q-1}_{q'=0}\nabla \cL(\mW^{t,q'}_m(\cS);\cS)}^2\right]\nonumber\\
        &\stackrel{(d)}{\leq}\eta^2C_0^2q\sum^M_{m=1}\sum^{q-1}_{q'=0}\E^t\E_{\cS}\left[\norm{\nabla \cL(\mW^{t,q'}_m(\cS^t),\cS^t) - \nabla \cL(\mW^{t,q'}_m(\cS),\cS)}^2\right.\nonumber\\
        &\qquad \left.+ \sum_{m'\neq m}\norm{\nabla \cL(\mW^{t,q'}_m(\cS);\cS)}^2\right]\nonumber\\
        & \stackrel{(e)}{=} \eta^2(M+1)C_0^2q\sum^M_{m=1}\sum^{q-1}_{q'=0}\E^t\E_{\cS}\norm{\nabla \cL(\mW^{t,q'}_m(\cS^t),\cS^t)}^2\nonumber\\
        & \stackrel{(f)}{=} \eta^2(M+1)C_0^2q\sum^{q-1}_{q'=0}\E^t\norm{\mG^{t,q'}}^2 \nonumber\\
        & \stackrel{(g)}{=} \eta^2(M+1)C_0^2q\sum^{q-1}_{q'=0}\E^t\left[\norm{\mG^{t,q'}-\E^t\mG^{t,q'}}^2 + \norm{\E^t \mG^{t,q'}}^2\right],\label{eq:term2}
    \end{align}}
where step $(a)$ uses Assumption~\ref{as:sample}, which states that $\cS$ is uniformly sampled from $\cD$, and applies Jensen's inequality, that is \begin{align*}
    &\norm{\E_{\cS}\nabla_{\mW_m}\cL(\mW^{t,q}(\cS^t);\cS) - \E_{\cS}\nabla \cL(\mW^{t,q}_{m}(\cS);\cS)}^2 \\
    &\leq \E_{\cS}\norm{\nabla_{\mW_m}\cL(\mW^{t,q}(\cS^t);\cS) - \nabla \cL(\mW^{t,q}_{m}(\cS);\cS)}^2;
\end{align*}
step $(b)$ applies \leref{le:smooth} and uses the fact that $\nabla \cL(\mW^{t,q}_m(\cS^t),\cS^t)$ is evaluated on $\mW^{t,q}_m(\cS^t)$ and $\mW^{t,0}_{-m}$; in step $(c)$ we expand the update steps until $t,0$ with \eqref{eq:update_S}; step $(d)$ applies Cauchy-Schwarz inequality; in step $(e)$ we reorder the sum and apply the i.i.d. Assumption~\ref{as:sample} to $\cS,\cS^t$; and in step $(g)$ we plug in the definition of $\mG$. This completes bounding the term $\E^t\norm{\nabla \cL(\mW^{t,q}, \cD)-\E^t\mG^{t,q}}^2$.

\subsubsection{Proof of the Main Result}\label{app:analysis:main}
Substituting the last term in \eqref{eq:descent} with \eqref{eq:term2}, we obtain that the following holds with probability $(1-\delta)^{Q}$:
\begingroup
\allowdisplaybreaks
\begin{align*}
    &\E^{t}[\cL(\mW^{t,q+1}, \cD)- \cL(\mW^{t,q}, \cD)] \leq -\frac{\eta}{2}\E^t\norm{\nabla\cL(\mW^{t,q}, \cD)}^2 -\frac{\eta}{2}(1-\eta C_0)\norm{\E^t\mG^{t,q}}^2 \nonumber\\
    & \quad + \eta^2C_0\E^t\norm{\mG^{t,q}-\E^t\mG^{t,q}}^2\nonumber\\
    & \quad + \frac{1+\eta C_0}{2C_0}\eta^2(M+1)C_0^2q\sum^{q-1}_{q'=0}\E^t\left[\norm{\mG^{t,q'}-\E^t\mG^{t,q'}}^2 + \norm{\E^t \mG^{t,q'}}^2\right] \\
    & \leq -\frac{\eta}{2}\E^t\norm{\nabla\cL(\mW^{t,q})}^2 -\frac{\eta}{2}(1-\eta L)\norm{\E^t\mG^{t,q}}^2 \\
    &\quad + \frac{1+\eta C_0}{2C_0}\eta^2(M+1)C_0^2q\sum^{q-1}_{q'=0}\E^t\norm{\E^t \mG^{t,q'}}^2\\
    & \quad + \eta^2C_0\cdot\left(1+\frac{(1+\eta C_0)\cdot (M+1)\cdot \eta Q^2}{2}\right)\cdot\frac{14M\sigma}{3},
\end{align*}
\endgroup
where in the second inequality, we set $\eta \leq \frac{1}{\sqrt{2}QC_0}$, plug in \eqref{eq:term1}, and use the fact that $q \leq Q$. Averaging over $t = 0,\dots, T-1$ and $q = 0,\dots, Q-1$ and reorganizing the terms, we obtain:
\begingroup
\allowdisplaybreaks
\begin{align*}
    &\frac{1}{TQ}\sum^{T-1}_{t=0}\sum^{Q-1}_{q=0}\E\norm{\nabla \cL(\mW^{t,q};\cD)}^2 \leq  \frac{2}{\eta TQ}\E[\cL(\mW^0)-\cL(\mW^{T,Q})] \\
    &-\frac{1-\eta C_0 \left(1+ (1+\eta C_0)\cdot (M+1)\cdot Q^2\right)}{TQ}\sum^{T-1}_{t=0}\sum^{Q-1}_{q=0}\E\norm{\E^t\mG^{t,q}}^2\\
    &+2\eta C_0\cdot\left(1+\frac{(1+\eta C_0)\cdot (M+1)\cdot \eta Q^2}{2}\right)\cdot\frac{14M\sigma}{3},
\end{align*}
\endgroup
which holds with probability at least $(1 - \delta)^{TQ}.$ Let $\delta = \delta'/TQ \in (0,1)$; then, the above equation holds with probability at least \[(1-\delta'/TQ)^{TQ}\geq 1- \delta'/TQ\times TQ = 1-\delta'.\]
Let \[1-\eta C_0 \left(1+ (1+\eta C_0)\cdot (M+1)\cdot Q^2\right)\geq0,\]
($\eta \leq \frac{1}{C_0\cdot(1+2MQ^2)}$) and apply Assumption~\ref{as:lower_bound}. Then, we have 
\begin{align}
    &\frac{1}{TQ}\sum^{T-1}_{t=0}\sum^{Q-1}_{q=0}\E\norm{\nabla \cL(\mW^{t,q};\cD)}^2 \leq  \frac{2(\cL(\mW^0)-\cL^\star)}{\eta TQ}+\frac{28\eta M\cdot\left(C_0 + \sqrt{M+1}Q\right)}{3}\sigma,
\end{align}
which holds with probability at least $1-\delta$, where 
\[\sigma = 64G_\ell^2L_f^2\log\left(\frac{2dTQ}{\delta}\right) + 128L_\ell^2\left(G_f^4+ \frac{1}{S}\right)\left(\log\left(\frac{2dTQ}{\delta}\right)+ \frac{1}{4}\right).\]
This completes the proof of \thref{thm:convergence}.

\subsection{Proof for \leref{le:smooth}}\label{app:proof:le}
In this subsection, we prove
\[\norm{\nabla_{\mW}\cL(\mW) - \nabla_{\mW'}\cL(\mW')} \leq C_0 \norm{\mW - \mW'}\]
and
\[\norm{\nabla_{\mW_m}\cL(\mW) - \nabla_{\mW'_m}\cL(\mW')} \leq C_0 \norm{\mW - \mW'}.\]

Note that $\nabla_{\mW_m}\cL(\mW)$ is a sub-vector of $\nabla \cL(\mW'),$ so $\norm{\nabla_{\mW_m}\cL(\mW) - \nabla_{\mW'_m}\cL(\mW')} \leq \norm{\nabla_{\mW}\cL(\mW) - \nabla_{\mW'}\cL(\mW')}.$ Therefore, we only need to prove the first inequality.

The gradient $\nabla\cL(\mW)$ can be expanded as 
\begin{equation}
    \begin{aligned}
        \nabla\cL(\mW) &= \nabla \ell(\yy, f_m(\mS , \mW))\\
        & = \nabla \ell(f_m(\mS,\mW)) \cdot \nabla_{\mW} f_m(\mS, \mW) = \nabla \ell(f_m) \cdot \nabla f_m(\cW),
    \end{aligned}
\end{equation}
where in the last equation we omit the irrelevant variables. Then, we have 
\begin{equation}
    \begin{aligned}
        &\Vert\nabla\cL(\mW) - \nabla\cL(\mW')\Vert = \norm{\nabla \ell(f_m) \cdot \nabla f_m(\mW) - \nabla \ell(f_m') \cdot \nabla f_m(\mW')}\\
        & = \norm{\nabla \ell(f_m) \cdot (\nabla f_m(\mW)-\nabla f_m(\mW')) + (\nabla \ell(f_m) - \nabla \ell(f_m')) \cdot \nabla f_m(\mW')}\\
        & \stackrel{(a)}{\leq}\norm{\nabla \ell(f_m) \cdot (\nabla f_m(\mW)-\nabla f_m(\mW'))} + \norm{(\nabla \ell(f_m) - \nabla \ell(f_m')) \cdot \nabla f_m(\mW')}\\
        & \stackrel{(b)}{\leq}\norm{\nabla \ell(f_m)} \norm{\nabla f_m(\mW)-\nabla f_m(\mW')} + \norm{\nabla \ell(f_m) - \nabla \ell(f_m')} \norm{\nabla f_m(\mW')}\\
        & \stackrel{(c)}{\leq}G_\ell L_f\norm{\mW-\mW'} + L_\ell\norm{f_m(\mW) - f_m(\mW')}\cdot G_f\\
        & \stackrel{\asref{as:smooth}}{\leq}(G_\ell L_f+ L_\ell G_f^2)\cdot \norm{\mW -\mW'},
    \end{aligned}
\end{equation}
where step $(a)$ uses the fact that $\norm{a+b}\leq \norm{a}+ \norm{b}$, step $(b)$ uses the fact that $\norm{ab}\leq\norm{a}\norm{b}$; and in step $(c)$ we apply \leref{le:smooth} (that is, for any $G$-smooth function $g$, its gradient is bounded as $\norm{\nabla g}\leq G$) to the first and the fourth terms and Lipschitz gradient to the second and the third terms. This completes the proof of \leref{le:smooth}.

\section{Experiment Details}\label{app:exp}

\subsection{Details of the Datasets}\label{app:exp:data}
\textbf{Planetoid}~\citep{yang2016revisiting}: This collection contains three citation datasets: Cora, PubMed, and CiteSeer. Each dataset contains one citation graph, where the nodes represent papers and edges represent citations. The node features are a bag of words and the classification target is the paper category. In the experiment, each client holds a non-overlapping block of node features and a subgraph that results from uniformly sampling 80\% of the edges.

\textbf{HeriGraph}~\citep{bai2022heri}: This collection contains three multi-modal graph datasets, each of which is constructed from heritage data posted on social media for a particular city (Suzhou, Amsterdam, and Venice). Each post contains user information, timestamp, geolocation, image, and text annotation. The posts are connected to form three subgraphs: a social subgraph, a spatial subgraph, and a temporal subgraph. The social subgraph is formed based on friendship and common-interest relations of the users. The spatial subgraph is formed based on the spatial proximity of the geolocations. The temporal subgraph is formed based on the temporal proximity of the posts. Each post has three blocks of image features and possibly text features; for classification, it belongs to one of nine heritage attributes. In the experiment, each client holds one of the three subgraphs and one of the three image feature blocks.

\textbf{Reddit}~\citep{hamilton2017inductive}: Reddit is a large online community where users post and comment on different topics. Each node represents a post and the features are the text of the post. Two posts are connected if the same user comments on both. The classification target is the community (subreddit) that a post belongs to. Similar to Planetoid, in the experiment, each client holds a non-overlapping block of node features and a subgraph that results from uniformly sampling 80\% of the edges.

\subsection{Details of the Hyperparameters}\label{app:exp:hyperparam}
Here we provide a list of hyperparameters for grid search in Table \ref{tab:parameters}. The optimal set of hyperparameters for each setting is tuned according to the range listed in the table.

\begin{table}[h]
  \centering
  \caption{Hyperparameter grid search range for the numerical experiments.}
  \label{tab:parameters}
  \vskip 5pt
  \small
  \begin{tabular}{c|l}
    \toprule
    Hyperparameter & Grid search range\\
    \midrule
    Hidden dimension of $\mH[l]$ & $\{128, 192, 256, 384\}$ \\
    Batch size $S$ & $\{16, 32\}$ \\
    Neighborhood sample size & $\{2,3,4,6,8\}$ \\
    Training rounds $T$ & $\{512, 640, 1024, 1152, 3200\}$\\
    Learning rate $\eta$ & $\{1,2,3.5,5,7,8\}\times\{10^{-1}, 10^{-2}, 10^{-3}\}$\\
    \bottomrule
  \end{tabular}
\end{table}

\end{document}